\documentclass{article}

\usepackage{amsmath}
\usepackage[dvipsnames]{xcolor}
\usepackage[parfill]{parskip}
\usepackage[numbers,sort&compress]{natbib}

\usepackage[T1]{fontenc}
\usepackage[tt=false, type1=true]{libertine}
\usepackage[varqu]{zi4}
\usepackage[libertine]{newtxmath}

\usepackage{bm}
\usepackage{booktabs}
\usepackage{array}
\usepackage{microtype}
\usepackage{enumitem}
\usepackage{graphicx}
\usepackage[font=small]{caption}
\usepackage{algorithm}
\usepackage{algorithmic}
\usepackage{url}
\usepackage{hyperref}
\definecolor{link}{HTML}{307ef1}
\hypersetup{
  colorlinks  = true,
  linkcolor   = link,
  citecolor   = link,
  urlcolor    = link,
  pdftitle    = {FlyWheel: Attention Kernels Built for the TPU},
  pdfkeywords = {TPU, Attention, Linear Attention, Gated DeltaNet, Pallas, LLM Inference},
}

\newcommand{\rpa}{\textit{RPA-v3}}
\newcommand{\splash}{\textit{Splash-Attention}}
\newcommand{\gdnv}{\textit{GDN-v3}}
\newcommand{\tpuinf}{\textit{tpu-inference}}
\newcommand{\tokamax}{\textit{Tokamax}}
\newcommand{\brpa}{\textit{Batched-RPA}}
\newcommand{\torchtpu}{\textit{vLLM TorchTPU}}

\newcommand{\github}{\raisebox{-1.3pt}{\includegraphics[height=1.05em]{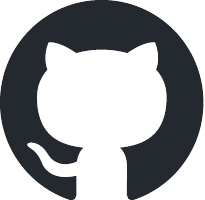}}}

\title{FlyWheel: Attention Kernels Built for the TPU}
\author{%
  Zishuo Bao$^{1,2}$ \quad Zeshen Zhang$^{1}$ \quad Linkun Chen$^{3}$ \quad Yucheng Lu$^{1,2}$ \\[0.6em]
  $^{1}$New York University \quad $^{2}$New York University Shanghai \quad $^{3}$Google \\[0.3em]
  \texttt{\{zb2466, yl5541\}@nyu.edu} \\[0.4em]
  \github\ \href{https://github.com/heavyball-research/flywheel-tpu}{\textcolor{link}{\textbf{Code}}}
}
\date{}

\begin{document}

\maketitle

\begin{abstract}
Attention kernels on Tensor Processing Units (TPUs) often underutilize their matrix units (MXUs) because of repeated matrix operand loads, vector-intensive computation, and dependencies that leave MXUs idle. We present FlyWheel, a set of fast TPU kernels for both softmax and linear attention. FlyWheel reformulates softmax and triangular inversion to reduce vector work, while coordinating data layouts to improve matrix operand reuse. It then overlaps independent computation to hide execution latency. We apply these techniques to both softmax attention and Gated DeltaNet on TPU v6e and v7x. FlyWheel speeds up softmax attention by up to $4.14\times$ over the state-of-the-art RPA kernel and $1.80\times$ over Splash Attention, reaching 81\% of peak BF16 throughput on v6e. It also speeds up Gated DeltaNet prefill by up to $3.61\times$. Integrated into vLLM serving Qwen3.8-27B, FlyWheel improves prefill throughput by up to $1.56\times$ while maintaining comparable accuracy. 
\end{abstract}

\section{Introduction}
\label{sec:intro}

Attention kernels are core to modern foundation models, and have been extensively optimized on GPUs. For instance, FlashAttention~\cite{dao2022flashattention,dao2024flashattention,shah2024flashattention,zadouri2026flashattention} and FlashInfer~\cite{ye2025flashinfer} accelerate softmax attention, while FLA~\cite{yang2024fla} provides efficient implementations of linear attention such as Gated DeltaNet (GDN)~\cite{yang2025gated}. These attention mechanisms are widely used in models such as Qwen~\cite{qwen38} and Kimi Linear~\cite{team2025kimi}, supporting applications from document reasoning to video understanding~\cite{hsieh2024ruler,kuratov2024babilong,fu2025video}. Much of this progress comes from designing algorithms around GPU hardware: reducing memory traffic, organizing work across threads, and overlapping matrix multiplication with other operations.

Tensor Processing Units (TPUs) offer another platform for serving these models, with recent benchmarks reporting competitive serving costs~\cite{semianalysis2026inferencex}. On GPUs, architecture-aware optimization has enabled FlashAttention-3 to reach 75\% of H100's peak FP16 throughput~\cite{shah2024flashattention}. Yet even state-of-the-art TPU attention kernels leave substantial performance unused. In our causal prefill benchmarks on TPU v6e, Ragged Paged Attention v3 (\rpa{})~\cite{jiang2026ragged} and Google's \splash{} kernel in \tokamax{}~\cite{tokamax} achieve only 18--53\% and 35--73\% of roofline throughput, respectively (\S\ref{sec:eval-softmax}). This gap raises a central question: 

\begin{center}
\textit{what limits attention performance on TPUs, and how can kernel designs better exploit their architecture?}
\end{center}

GPU attention optimizations do not transfer directly to TPUs because the architectures differ in three key respects. First, TPU layout constraints and weight-stationary matrix units (MXUs) make data placement and operand reuse central to matrix efficiency. Layout choices determine tile utilization, while repeated operand loads can limit throughput even when data is already on chip. Second, TPUs provide limited vector throughput relative to matrix compute, making softmax reductions, rescaling, and other elementwise operations potential bottlenecks. Reducing this work also requires managing register pressure: in our v6e experiments, the additional state required by FlashAttention-4's conditional rescaling~\cite{zadouri2026flashattention} offsets its savings in vector work (\S\ref{sec:disc-softmax}). Third, a TPU TensorCore exposes a single-threaded programming model with compiler-managed overlap across execution units, rather than GPU-style warp scheduling. FlashAttention-3~\cite{shah2024flashattention} combines asynchronous matrix operations with inter- and intra-warpgroup pipelining. On TPUs, kernels must instead expose independent work in a form that the compiler can schedule across the matrix and vector units.

To address these gaps,
we present FlyWheel: a suite of TPU-native kernels for softmax attention and Gated DeltaNet, written in Pallas~\cite{pallas}. FlyWheel coordinates layouts across operator boundaries so that projections produce and consume the attention kernel's layout directly, and handles ragged sequence boundaries on chip. It amortizes MXU operand loads by increasing reuse and combining matrix products that share an operand. To reduce pressure on the vector units, FlyWheel removes repeated softmax rescaling with a checked static anchor and computes the GDN triangular inverse with eight BF16 matrix multiplications. Finally, it interleaves independent work across attention fragments and GDN chunks to hide latency and overlap matrix and vector computation. Together, these designs reduce the overhead surrounding matrix multiplication and keep the MXUs busy (\S\ref{sec:design}).

We evaluate FlyWheel on TPU v6e and v7x (\S\ref{sec:eval}). FlyWheel accelerates softmax attention by up to $4.14\times$ over \rpa{} and $1.80\times$ over \splash{}, reaching 81\% of peak BF16 throughput on v6e. It accelerates GDN prefill by up to $3.61\times$ over the state-of-the-art \gdnv{} kernel. These kernel improvements translate into application-level gains: in vLLM serving Qwen3.8-27B at 512K tokens, FlyWheel improves prefill throughput by $1.53\times$ on RULER and $1.56\times$ on BABILong on v6e, and by up to $1.35\times$ on v7x. On Video-MME, it accelerates language model prefill by up to $1.29\times$ and $1.24\times$ on the two generations, respectively, while maintaining comparable benchmark accuracy (\S\ref{sec:eval-e2e}).

Our contributions in this paper can be summarized as follows:
\begin{itemize}[nosep, leftmargin=12pt]
\item We identify shared architectural bottlenecks in TPU softmax and linear attention and show why established GPU optimizations do not directly address them.
\item We develop FlyWheel, a suite of TPU-native attention kernels that combines coordinated layouts, matrix operand reuse, reduced vector work, and explicit pipelining.
\item We demonstrate kernel speedups on TPU v6e and v7x and translate them into application-level gains in vLLM: up to $1.53\times$ and $1.56\times$ higher prefill throughput on RULER and BABILong, respectively, and $1.29\times$ faster language model prefill on Video-MME, while maintaining comparable accuracy.
\end{itemize}

\section{Background}
\label{sec:background}

We first describe the computations underlying softmax attention and Gated DeltaNet, then examine how they interact with TPU hardware. Throughout the paper, our discussion focuses on prefill kernels\footnote{We defer the discussion of decode kernels to Appendix~\ref{app:decode}.} on TPU v6e and v7x; GPU comparisons refer to NVIDIA Hopper and later architectures.

\subsection{Softmax Attention}
\label{sec:bg-softmax}

Given queries, keys, and values $\bm{Q},\bm{K},\bm{V} \in \mathbb{R}^{T \times D}$, where $T$ is the sequence length and $D$ is the head dimension, softmax attention~\cite{vaswani2017attention} computes
\begin{equation}
\bm{P} = \mathrm{softmax}(\bm{Q}\bm{K}^\top), \qquad \bm{O} = \bm{P}\bm{V},
\label{eq:attn}
\end{equation}
where $\bm{P} \in \mathbb{R}^{T \times T}$ and $\bm{O} \in \mathbb{R}^{T \times D}$. Softmax is applied row-wise. We omit the scaling factor and attention mask for simplicity. The computation is applied independently to each query head; grouped-query attention shares keys and values across query heads~\cite{ainslie2023gqa}.

\textbf{Blockwise computation.}
FlashAttention~\cite{dao2022flashattention} avoids materializing $\bm{P}$ in HBM by processing keys and values in blocks. For each query row, online softmax~\cite{milakov2018online} maintains a scalar running maximum $m$, a scalar normalization sum $l$, and an output accumulator $\bm{o} \in \mathbb{R}^{1 \times D}$. For a block of $B$ keys, let $\bm{s} \in \mathbb{R}^{1 \times B}$ be the scores and $\bm{V}_b \in \mathbb{R}^{B \times D}$ the values. The update is
\begin{equation}
\begin{aligned}
m' &= \max(m,\max_j s_j), \qquad \bm{p}=e^{\bm{s}-m'}, \\
l &\leftarrow r l+\textstyle\sum_j p_j, \\
\bm{o} &\leftarrow r \bm{o}+\bm{p}\bm{V}_b,
\end{aligned}
\label{eq:online}
\end{equation}
where $r=e^{m-m'}$ rescales the previous accumulators. The algorithm then sets $m \leftarrow m'$ and returns $\bm{o}/l$ after the final block.

Each block contains two matrix multiplications separated by vector operations: maximum reduction, exponentiation, sum reduction, and accumulator rescaling. The running maximum stabilizes the exponentials but also introduces rescaling dependencies across blocks. These vector operations and dependencies can limit matrix throughput. Appendix~\ref{app:reference-algorithms} provides the online-softmax reference in Algorithm~\ref{alg:fa-reference}.

\subsection{Linear Attention}
\label{sec:bg-linear}

Unlike softmax attention, which attends explicitly to previous keys and values, linear attention~\cite{katharopoulos2020transformers} summarizes the token history in a fixed-size state matrix. We focus on Gated DeltaNet (GDN)~\cite{yang2025gated}, which updates this state using the delta rule~\cite{schlag2021linear}. Let $\bm{q}_t,\bm{k}_t\in\mathbb{R}^{1\times d_k}$ and $\bm{v}_t\in\mathbb{R}^{1\times d_v}$ denote the query, key, and value rows for token $t$, where $d_k$ and $d_v$ are the query/key and value dimensions, respectively. The state $\bm{S}_t\in\mathbb{R}^{d_k\times d_v}$ and output $\bm{o}_t\in\mathbb{R}^{1\times d_v}$ satisfy
\begin{equation}
\begin{aligned}
\bm{S}_t &= \bm{S}_{t-1}
 + \beta_t \bm{k}_t^\top(\bm{v}_t-\bm{k}_t\bm{S}_{t-1}), \\
\bm{o}_t &= \bm{q}_t\bm{S}_t,
\end{aligned}
\label{eq:gdn}
\end{equation}
where the scalar $\beta_t$ controls the update magnitude. The residual $\bm{v}_t-\bm{k}_t\bm{S}_{t-1}$ corrects the state's prediction of the current value. For clarity, we omit GDN's decay gate, which scales the previous state before the update.

\textbf{Chunked computation.}
In GDN, prefill kernels often group tokens into chunks to express the recurrence as matrix products~\cite{yang2024parallelizing,yang2025gated}. Within a chunk of $C$ tokens, we use local indices $t=1,\ldots,C$ and stack the query, key, and value rows $\bm{q}_t$, $\bm{k}_t$, and $\bm{v}_t$ into $\bm{Q},\bm{K}\in\mathbb{R}^{C\times d_k}$ and $\bm{V}\in\mathbb{R}^{C\times d_v}$, respectively. Let $\bm{\beta}=\mathrm{diag}(\beta_1,\ldots,\beta_C)$ collect their update coefficients. Define
\[
\bm{A}_{kk}=\mathrm{tril}_{-1}(\bm{\beta}\bm{K}\bm{K}^\top),
\qquad
\bm{A}_{qk}=\mathrm{tril}_{0}(\bm{Q}\bm{K}^\top),
\]
both of shape $C\times C$, where $\mathrm{tril}_{-1}$ and $\mathrm{tril}_{0}$ retain the lower triangle without and with the diagonal, respectively. Given the state $\bm{S}\in\mathbb{R}^{d_k\times d_v}$ entering the chunk, the computation is
\begin{equation}
\begin{aligned}
\bm{V}_{\mathrm{new}} &= (\bm{I}+\bm{A}_{kk})^{-1}\bm{\beta}(\bm{V}-\bm{K}\bm{S}), \\
\bm{O} &= \bm{Q}\bm{S}+\bm{A}_{qk}\bm{V}_{\mathrm{new}}, \\
\bm{S} &\leftarrow \bm{S}+\bm{K}^\top\bm{V}_{\mathrm{new}},
\end{aligned}
\label{eq:chunk}
\end{equation}
where $\bm{I}$ is the $C\times C$ identity matrix, $\bm{V}_{\mathrm{new}}\in\mathbb{R}^{C\times d_v}$ contains the value updates, and $\bm{O}\in\mathbb{R}^{C\times d_v}$ stacks the output rows $\bm{o}_t$. As above, decay factors are omitted; they introduce additional scaling without changing the matrix dimensions~\cite{yang2025gated}.

The matrices $\bm{A}_{kk}$, $\bm{A}_{qk}$, and $(\bm{I}+\bm{A}_{kk})^{-1}$ depend only on chunk-local inputs and can be computed independently across chunks. In contrast, $\bm{V}_{\mathrm{new}}$, $\bm{O}$, and the state update depend on the incoming state and must respect the recurrence. The computation also exposes operand reuse: constructing $\bm{A}_{kk}$ and $\bm{A}_{qk}$ uses the same $\bm{K}^\top$, while the intra-chunk output and state increment share $\bm{V}_{\mathrm{new}}$. Appendix~\ref{app:reference-algorithms} gives a reference algorithm using forward substitution.

\textbf{Additional layer operations.}
The query, key, and value vectors entering the recurrence are obtained through depthwise causal convolution and SiLU, with additional per-head L2 normalization for queries and keys. GDN prefill thus combines these preprocessing operations with small matrix products and triangular inversion, motivating joint optimization of the full computation.

\subsection{TPU Architecture and Attention Bottlenecks}
\label{sec:bg-diff}

A TPU v6e chip contains one TensorCore, while a v7x chip contains two~\cite{jouppi2026google}. Each TensorCore uses a VLIW instruction stream to coordinate matrix, vector, and specialized execution units~\cite{jouppi2020domain}, with DMA engines transferring data between HBM and on-chip vector memory (VMEM). These units can execute concurrently when dependencies permit. Layout overhead, non-matrix computation, and execution stalls also affect GPU attention; the challenge is to address them under the TPU's data organization and execution model.

\textbf{Data layout and matrix operand reuse.}
TPU vector registers contain $8\times128$ 32-bit words, organized as sublanes and lanes; packing two BF16 elements per word accommodates a $16\times128$ tile. Layout choices therefore affect padding, register utilization, and the shapes available to matrix operations. The MXUs use weight-stationary systolic arrays~\cite{jouppi2026google}: for $\bm{X}\bm{W}$, the stationary operand $\bm{W}$ is loaded into the array and reused as rows of $\bm{X}$ stream through it. This loading cost must be amortized even when operands already reside on chip.

For softmax attention, the layout determines how many query rows fit in a tile and reuse each key or value load. Tiling tokens also introduces alignment constraints at ragged sequence boundaries. For GDN, small chunk matrices offer less reuse per load, although several products share the same right operand (Eq.~\ref{eq:chunk}). Its input layout must also support both convolution and per-head matrix computation. Efficient layouts must therefore support operand reuse without introducing costly conversions between operators.

\textbf{Non-matrix work on the critical path.}
TPUs provide much higher throughput for matrix multiplication than for vector operations~\cite{jouppi2020domain}. Softmax attention places reductions, exponentials, and accumulator rescaling between its two matrix products. In GDN, forward substitution introduces dependent row updates, while convolution and normalization add vector arithmetic and data rearrangement. Consequently, reducing matrix execution time alone may leave overall kernel latency largely unchanged. This imbalance also motivates GPU optimizations such as FlashAttention-4~\cite{zadouri2026flashattention}, but their benefit on TPUs depends on the resulting instruction mix, register requirements, and dependencies.

\textbf{Exposing work for compiler scheduling.}
FlashAttention-3~\cite{shah2024flashattention} exploits asynchronous matrix operations and inter- and intra-warpgroup pipelining. A TPU TensorCore instead relies on compiler scheduling within a single instruction stream. Concurrent execution is supported, but sufficient independent work must be visible to the compiler.

In softmax attention, each fragment contains a score-to-softmax-to-output dependency chain, and online rescaling couples successive fragments through their accumulators. In GDN, triangular inversion contains dependent operations, while the recurrent state links successive chunks. However, other softmax fragments and GDN's state-independent chunk computations provide work that can overlap these chains. Kernel structure and loop boundaries determine whether the compiler can exploit this independence.

These observations guide the two kernel designs in \S\ref{sec:design}. For softmax attention, FlyWheel combines larger per-head tiles with a checked static anchor and fragment pipelining (\S\ref{sec:design-softmax}). For GDN, it combines shared-operand products with matrix-based inversion and cross-chunk pipelining (\S\ref{sec:design-gdn}).


\section{FlyWheel}
\label{sec:design}

FlyWheel addresses the bottlenecks in \S\ref{sec:bg-diff} by connecting data layout, computational reformulation, and scheduling. In softmax attention, per-head tiles increase operand reuse, while a checked static anchor simplifies the dependencies between fragments. In GDN, shared-operand products and hierarchical inversion move more work onto the MXU, and cross-chunk scheduling hides the resulting matrix dependencies.

\subsection{Softmax Attention}
\label{sec:design-softmax}

\begin{figure}[t]
\centering
\includegraphics[width=0.65\linewidth]{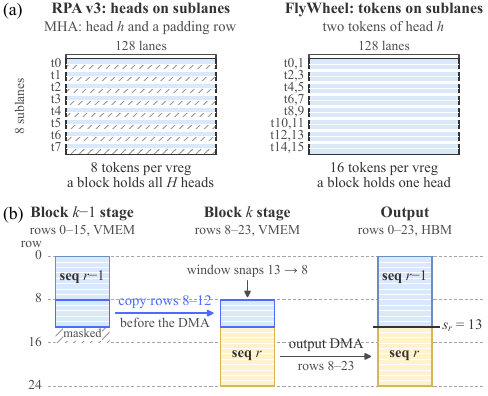}
\caption{Per-head tiling with ragged boundaries. (a) Under MHA, \rpa{} pads each query head and holds 8 tokens per BF16 register; FlyWheel holds 16 tokens of one head without padding. (b) A sequence starting at row 13 is written through a window aligned at row 8; FlyWheel first restores rows 8--12 from the previous VMEM buffer.}
\label{fig:layout}
\end{figure}

\textbf{Layout and operand reuse.}
Attention inputs have logical shapes $\bm{Q}\in\mathbb{R}^{T\times H_q\times D}$ and $\bm{K},\bm{V}\in\mathbb{R}^{T\times H_{kv}\times D}$, where $T$ is the token count, $H_q$ and $H_{kv}$ are the query and key/value head counts, and $D$ is the head dimension. Mapping these axes to registers determines how much computation can reuse each loaded key or value tile. In \rpa{}~\cite{jiang2026ragged}, a query tile spans all heads, so its accumulator storage grows with the head count and restricts the number of query rows. Under MHA, packing heads on sublanes also leaves half of each BF16 register as padding (Figure~\ref{fig:layout}a).

FlyWheel instead uses a head-major layout, packing two tokens from the same head into each sublane. This removes the MHA padding and decouples query tile size from head count: a tile holds 256 query rows, allowing each stationary key or value load to serve 256 rows. GQA uses the same organization, with each query head accessing its shared key/value head.

Token tiling introduces a second problem: ragged sequences need not start at DMA-aligned rows. Figure~\ref{fig:layout}b shows a sequence starting at row 13 whose DMA window begins at row 8. FlyWheel preserves rows 8--12 by copying their previously computed outputs from the preceding VMEM buffer before writing the aligned window. Together with projections that directly produce and consume the kernel's layout, this allows larger per-head tiles without separate HBM layout conversions or output gathers.

\textbf{Reducing vector work with a checked static anchor.}
The layout improves matrix reuse, but each online-softmax step still inserts a maximum reduction and accumulator rescaling between matrix products. FlyWheel removes this repeated work by using a fixed reference for the exponential weights. Following Eq.~\ref{eq:online}, it sets a scalar anchor $a$ to the maximum score in each query row's first key/value tile and accumulates
\begin{equation}
\begin{aligned}
\bm{p} &= e^{\bm{s}-a}, \\
l &\leftarrow l+\textstyle\sum_j p_j, \\
\bm{o} &\leftarrow \bm{o}+\bm{p}\bm{V}_b.
\end{aligned}
\label{eq:anchor}
\end{equation}
The shapes are unchanged from the Background: $\bm{s},\bm{p}\in\mathbb{R}^{1\times B}$, $\bm{V}_b\in\mathbb{R}^{B\times D}$, and $\bm{o}\in\mathbb{R}^{1\times D}$, with scalar $l$. The final division $\bm{o}/l$ cancels the common factor $e^{-a}$ in exact arithmetic.

The anchor need not remain the maximum, but the accumulated values must remain within numerical range. Since the weights are nonnegative, their sum $l$ bounds every weight. FlyWheel checks $l\leq\Lambda$ after each key/value block. The comparison also fails on NaN and $+\infty$, and $|\bm{o}|\leq l\max|\bm{V}|$ elementwise, so $\bm{o}$ needs no separate check. If any row fails, FlyWheel replays that block with guarded anchor updates in the same launch (Algorithm~\ref{alg:guarded-replay}). Unlike FlashAttention-4~\cite{zadouri2026flashattention}, which checks each fragment's maximum, FlyWheel checks the already-accumulated $l$, saving the per-fragment reduction, the rescaling branch, and their vector registers.

\begin{figure}[t]
\centering
\includegraphics[width=0.65\linewidth]{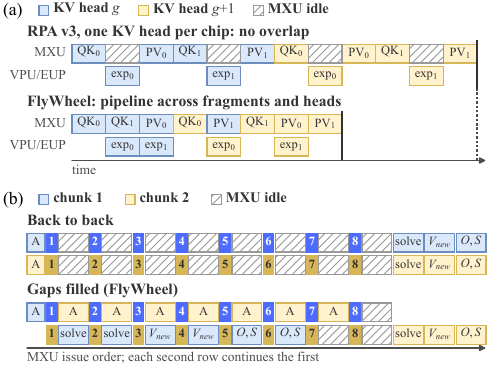}
\caption{Filling dependency-induced execution gaps. (a) \rpa{} leaves the MXU idle during exponentiation; FlyWheel overlaps each fragment's scores and exponentials with the previous fragment's PV product. (b) Gaps between the dependent inverse products 1--8 of one GDN chunk are filled with the other chunk's stacked product (A) and state-dependent work. Lengths are schematic.}
\label{fig:schedule}
\end{figure}

\textbf{Exposing overlap across fragments.}
A fragment pairs one query tile with one key/value tile. Its QK matmul and softmax do not depend on the preceding fragment's PV matmul; only the output accumulation is sequential. This independence also holds under online softmax with a running max, and FlashAttention-3 uses it to overlap softmax with GEMMs on GPUs~\cite{shah2024flashattention}. FlyWheel software-pipelines fragments at distance one. It defers each fragment's PV matmul and issues it after the next fragment's QK matmul and exponentials. It also keeps these fragments in one compiler basic block, so the scheduler can overlap one fragment's MXU work with the next fragment's vector work. Figure~\ref{fig:schedule}a contrasts this schedule with the gaps left by \rpa{} when only one key/value head is available per device. The overlapped work comes from consecutive fragments of the same head, so FlyWheel keeps the MXU busy even within a single head. The pipeline also continues across head boundaries: the first QK matmul of one head overlaps the last PV matmul of the previous head. The fixed anchor still makes each pipelined step cheaper. No fragment rescales the output accumulator, and each fragment's exponentials depend only on its own scores.

Algorithm~\ref{alg:FlyWheel-softmax} summarizes this schedule for one head. It batches the row-wise notation: $\bm{s},\bm{p}\in\mathbb{R}^{B_q\times B_k}$, $\bm{a},\bm{l}\in\mathbb{R}^{B_q}$, and $\bm{o}\in\mathbb{R}^{B_q\times D}$, with row-wise broadcasting. 
The preceding weights are retained until their output product completes, and the last product drains the pipeline.

\begin{algorithm}[t]
\caption{FlyWheel softmax attention}
\label{alg:FlyWheel-softmax}
\small
\begin{algorithmic}[1]
\REQUIRE $\bm{Q},\bm{K},\bm{V}\in\mathbb{R}^{T\times D}$; range threshold $\Lambda$
\ENSURE $\bm{O}\in\mathbb{R}^{T\times D}$
\FOR{each query tile $\bm{Q}_i$}
    \STATE Partition keys and values into $(\bm{K}_j,\bm{V}_j)_{j=1}^{J}$, grouped into blocks
    \STATE $\bm{s}\leftarrow\textsc{Scores}(\bm{Q}_i,\bm{K}_1)$
    \STATE $\bm{a}\leftarrow\operatorname{rowmax}(\bm{s})$
    \STATE $\bm{p}\leftarrow e^{\bm{s}-\bm{a}}$; $\bm{l}\leftarrow\operatorname{rowsum}(\bm{p})$; $\bm{o}\leftarrow\bm{0}$
    \FOR{$j=2,\ldots,J$}
        \STATE \textbf{Interleave:}
        \STATE \quad $\bm{s}\leftarrow\textsc{Scores}(\bm{Q}_i,\bm{K}_j)$
        \STATE \quad $\bm{o}\leftarrow\bm{o}+\bm{p}\bm{V}_{j-1}$
        \STATE \quad $\bm{p}_{\mathrm{next}}\leftarrow e^{\bm{s}-\bm{a}}$
        \STATE $\bm{p}\leftarrow\bm{p}_{\mathrm{next}}$
        \STATE $\bm{l}\leftarrow\bm{l}+\operatorname{rowsum}(\bm{p})$
        \IF{$\bm{K}_j$ ends a block $\beta$ and not ($\bm{l}\leq\Lambda$ for all rows)}
            \STATE $(\bm{a},\bm{l},\bm{o})\leftarrow\textsc{GuardedReplay}(\bm{Q}_i,\beta)$; $\bm{p}\leftarrow\bm{0}$
        \ENDIF
    \ENDFOR
    \STATE $\bm{o}\leftarrow\bm{o}+\bm{p}\bm{V}_J$
    \STATE $\bm{O}_i\leftarrow\bm{o}/\bm{l}$
\ENDFOR
\STATE \textbf{return} $\bm{O}$
\end{algorithmic}
\end{algorithm}

\textbf{Guarded replay.}
Algorithm~\ref{alg:guarded-replay} specifies the fallback used by Algorithm~\ref{alg:FlyWheel-softmax}. It recomputes only the key/value block that failed the check. The kernel keeps the anchor and accumulators on entry to each block in a second buffer, so the replay starts from them rather than from the first key tile; the first block enters with the anchor from $\bm{K}_1$ and zero accumulators. The entry state of a block holds the contributions of all fragments before it: the preceding block's last output product is issued in the block's first step, before the block's check. The replay checks each fragment before exponentiation. It also issues the block's last output product, so Algorithm~\ref{alg:FlyWheel-softmax} clears the pending weights $\bm{p}$ after the replay. The anchor is updated independently for each row only when the fragment maximum exceeds the row's anchor by more than a score margin $\tau\geq0$. All exponentials are written in natural units, consistently with Eq.~\ref{eq:anchor}; the implementation converts the margin to base-two units when using \texttt{exp2}.

\begin{algorithm}[htbp]
\caption{\textsc{GuardedReplay}: checked softmax recomputation}
\label{alg:guarded-replay}
\small
\begin{algorithmic}[1]
\REQUIRE Query tile $\bm{Q}_i$; block $\beta$ of key/value tiles $(\bm{K}_j,\bm{V}_j)_{j\in\beta}$; margin $\tau\geq0$
\ENSURE Anchor and accumulators $(\bm{a},\bm{l},\bm{o})$ after $\beta$
\STATE $(\bm{a},\bm{l},\bm{o})\leftarrow$ anchor and accumulators on entry to $\beta$
\FOR{each $j\in\beta$ in order}
    \STATE $\bm{s}\leftarrow\textsc{Scores}(\bm{Q}_i,\bm{K}_j)$
    \STATE $\bm{m}\leftarrow\operatorname{rowmax}(\bm{s})$
    \STATE $\bm{a}'\leftarrow\operatorname{where}(\bm{m}-\bm{a}>\tau,\bm{m},\bm{a})$
    \STATE $\bm{r}\leftarrow e^{\bm{a}-\bm{a}'}$
    \STATE $\bm{l}\leftarrow \bm{r}\odot \bm{l}$; $\bm{o}\leftarrow \bm{r}\odot \bm{o}$
    \STATE $\bm{a}\leftarrow \bm{a}'$; $\bm{p}\leftarrow e^{\bm{s}-\bm{a}}$
    \STATE $\bm{l}\leftarrow \bm{l}+\operatorname{rowsum}(\bm{p})$
    \STATE $\bm{o}\leftarrow \bm{o}+\bm{p}\bm{V}_j$
\ENDFOR
\STATE \textbf{return} $(\bm{a},\bm{l},\bm{o})$
\end{algorithmic}
\end{algorithm}

\textbf{Anchor updates.}
The shapes and broadcasting conventions are the same as in Algorithm~\ref{alg:FlyWheel-softmax}. The selection in $\operatorname{where}$ is row-wise, and $\bm{r}$ scales each row of $\bm{o}$. For a row whose anchor is unchanged, the rescaling factor is 1 and its rescaling can be omitted. If the anchor changes from $\bm{a}$ to $\bm{a}'$, multiplying both accumulators by $e^{\bm{a}-\bm{a}'}$ expresses all previous contributions relative to the new anchor. Thus, in exact arithmetic, their ratio remains the softmax output. As in the main algorithm, the first key tile must contain a valid key for every query row; fully masked later fragments contribute zero.

\textbf{Choosing the range margin.}
Let $V_{\max}$ bound the absolute entries of $\bm{V}$, and let $F$ be a conservative upper limit supported by both the exponential operands and the FP32 accumulators. One sufficient range condition, before accounting for rounding, is
\begin{equation}
T e^{\tau}\max(1,V_{\max})\leq F e^{-\delta}.
\label{eq:guarded-range}
\end{equation}
Here the sequence length $T$ bounds the number of keys of each query row. After every guarded update, each score is at most
 $\tau$ above its row's anchor, so each weight is at most $e^{\tau}$, $l\leq Te^{\tau}$, and $|\bm{o}|\leq Te^{\tau}V_{\max}$; Eq.~\ref{eq:guarded-range} keeps both bounds below $F$. The margin $\delta\geq0$ reserves headroom below $F$ for rounding.
For the static-anchor pass, a conservative acceptance bound is $\Lambda=e^{\tau}$, since $\max_j p_j\leq l$ for each row. The comparison $\bm{l}\leq\Lambda$ in Algorithm~\ref{alg:FlyWheel-softmax} is false for NaN and $+\infty$, so a nonfinite sum also triggers a replay of the failing block. Since $|\bm{o}|\leq \bm{l}V_{max}$ elementwise, an accepted $\bm{l}$ also keeps $\bm{o}$ within range, and $\bm{o}$ needs no separate check. The concrete threshold and numerical headroom must also account for the implementation's operand casts and reductions.

\subsection{Gated DeltaNet}
\label{sec:design-gdn}

\textbf{Layout and operand reuse.}
GDN organizes computation into small recurrent chunks, limiting the rows available to reuse each operand. FlyWheel therefore reuses operands across products within each chunk. It consumes the projection's BF16 output directly, avoiding the \gdnv{} wrapper's HBM upcast and retiling, and stacks the products in Eq.~\ref{eq:chunk} that share a right operand:
\begin{equation}
\begin{aligned}
\begin{bmatrix}
\bm{\beta}\bm{K}\\
\bm{Q}
\end{bmatrix}\bm{K}^\top
&=
\begin{bmatrix}
\bm{\beta}\bm{K}\bm{K}^\top\\
\bm{Q}\bm{K}^\top
\end{bmatrix},\\
\begin{bmatrix}
\bm{K}^\top\\
\bm{A}_{qk}
\end{bmatrix}\bm{V}_{\mathrm{new}}
&=
\begin{bmatrix}
\bm{K}^\top\bm{V}_{\mathrm{new}}\\
\bm{A}_{qk}\bm{V}_{\mathrm{new}}
\end{bmatrix}.
\end{aligned}
\label{eq:gdn-stack}
\end{equation}
The first product loads $\bm{K}^\top$ once to construct both $\bm{A}_{kk}$ and $\bm{A}_{qk}$ after triangular masking. The second reuses $\bm{V}_{\mathrm{new}}$ for the state increment and intra-chunk output. Thus, four separate matrix products become two larger products with shared operand loads.

\begin{figure}[t]
\centering
\includegraphics[width=0.65\linewidth]{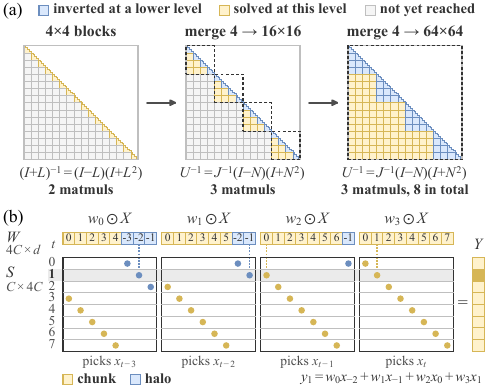}
\caption{Replacing vector operations with matrix products. (a) Hierarchical triangular inversion merges $4\times4$ blocks into $16\times16$ and then $64\times64$ blocks. (b) A four-tap causal convolution ($C=8$) as one product $\bm{Y}=\bm{S}\bm{W}$, where $\bm{W}$ stacks the tap-scaled inputs and halo rows, and the constant selector $\bm{S}$ ($\bm{S}_{\mathrm{conv}}$ in the text) sums the required rows.}
\label{fig:vector}
\end{figure}

\textbf{Reducing vector work with hierarchical inversion.}
Operand reuse does not remove the triangular solve on GDN's critical path. FlyWheel replaces its dependent forward-substitution updates with a matrix-based inverse of $\bm{U}=\bm{I}+\bm{A}_{kk}\in\mathbb{R}^{C\times C}$. A polynomial inverse applied to the entire chunk can produce large intermediate powers and cancellation~\cite{sobczyk2026fast}. Following the hierarchical approach of prior work, which merges pairs of blocks using FP32 products~\cite{sobczyk2026fast}, FlyWheel instead inverts $4\times4$ blocks and merges four at a time using BF16 operands with FP32 accumulation. Figure~\ref{fig:vector}a shows this progression: each stage reuses the diagonal blocks already inverted in blue and processes the blocks shown in orange.

\begin{algorithm}[t]
\caption{FlyWheel Gated DeltaNet}
\label{alg:FlyWheel-gdn}
\small
\begin{algorithmic}[1]
\REQUIRE $\bm{Q},\bm{K}\in\mathbb{R}^{T\times d_k}$, $\bm{V}\in\mathbb{R}^{T\times d_v}$; coefficients $\beta_t$; state $\bm{S}\in\mathbb{R}^{d_k\times d_v}$
\ENSURE $\bm{O}\in\mathbb{R}^{T\times d_v}$ and updated $\bm{S}$

\STATE Partition inputs into $(\bm{Q}_c,\bm{K}_c,\bm{V}_c,\bm{\beta}_c)_{c=1}^{N}$ of $C=64$ rows
\STATE $[\bm{R}_{kk,1};\bm{R}_{qk,1}]\leftarrow[\bm{\beta}_1\bm{K}_1;\bm{Q}_1]\bm{K}_1^\top$
\FOR{$c=1,\ldots,N+1$}
    \STATE \textbf{Interleave} (let $p=c-1$; skip chunks outside $1..N$):
    \STATE \quad $\bm{U}_c^{-1}\leftarrow\textsc{Inverse4}(\bm{I}+\operatorname{tril}_{-1}(\bm{R}_{kk,c}))$
    \STATE \quad $[\bm{R}_{kk,c+1};\bm{R}_{qk,c+1}]\leftarrow[\bm{\beta}_{c+1}\bm{K}_{c+1};\bm{Q}_{c+1}]\bm{K}_{c+1}^\top$
    \STATE \quad $[\tilde{\bm{V}}_p\,|\,\bm{W}_p]\leftarrow\bm{U}_p^{-1}\bm{\beta}_p[\bm{V}_p\,|\,\bm{K}_p]$
    \STATE \quad $[\bm{P}_p;\bm{O}_{\mathrm{inter},p}]\leftarrow[\bm{W}_p;\bm{Q}_p]\bm{S}$
    \STATE \quad $\bm{V}_{\mathrm{new},p}\leftarrow\tilde{\bm{V}}_p-\bm{P}_p$
    \STATE \quad $[\Delta\bm{S}_p;\bm{O}_{\mathrm{intra},p}]\leftarrow[\bm{K}_p^\top;\operatorname{tril}_0(\bm{R}_{qk,p})]\bm{V}_{\mathrm{new},p}$
    \STATE \quad $\bm{O}_p\leftarrow\bm{O}_{\mathrm{inter},p}+\bm{O}_{\mathrm{intra},p}$;\ $\bm{S}\leftarrow\bm{S}+\Delta\bm{S}_p$
\ENDFOR

\STATE \textbf{return} $\bm{O},\bm{S}$
\end{algorithmic}
\end{algorithm}

At each merge, write $\bm{U}=\bm{J}+\bm{E}$, where $\bm{J}$ contains the four diagonal blocks and $\bm{E}$ the strictly lower off-diagonal blocks. The preceding level supplies $\bm{J}^{-1}$. Defining $\bm{N}=\bm{E}\bm{J}^{-1}$ gives $\bm{N}^4=0$ and
\begin{equation}
\bm{U}^{-1}=\bm{J}^{-1}(\bm{I}-\bm{N})(\bm{I}+\bm{N}^2).
\label{eq:tinv-block}
\end{equation}
Because $\bm{J}^{-1}\bm{N}$ and $\bm{N}^2$ share the right operand $\bm{N}$, one stacked product forms both. Each merge therefore takes three products: $\bm{E}\bm{J}^{-1}$, the stacked product, and the product with $\bm{I}+\bm{N}^2$.

The hierarchy limits the polynomial to a four-block system at each level, rather than forming high powers of the full-chunk matrix. For $C=64$, two merge levels complete the inverse.

Algorithm~\ref{alg:inverse4} gives the complete \textsc{Inverse4} routine used by Algorithm~\ref{alg:FlyWheel-gdn}. Its input is $\bm{U}=\bm{I}+\bm{A}_{kk}\in\mathbb{R}^{64\times64}$. It first inverts the independent $4\times4$ diagonal blocks: for a block $\bm{I}+\bm{L}$ with strictly lower-triangular $\bm{L}$, $\bm{L}^4=0$ and $(\bm{I}+\bm{L})^{-1}=(\bm{I}-\bm{L})(\bm{I}+\bm{L}^2)$, which takes two products: $\bm{W}=\bm{L}\bm{L}$ and $(\bm{I}-\bm{L})(\bm{I}+\bm{W})$. Each merge level then processes consecutive index blocks $b$ of $h\in\{16,64\}$ rows, and $\bm{U}_{b,b}$ denotes the diagonal block of $\bm{U}$ on $b$. Its four diagonal subblocks form $\bm{J}$, and $\bm{R}_b$ collects their inverses from the preceding level, so $ \bm{R}_b=\bm{J}^{-1}$ in exact arithmetic. The matrix $\bm{N}=\bm{E}\bm{J}^{-1}$ is strictly lower triangular with respect to four blocks, hence $\bm{N}^4=0$. Since $\bm{U}_{b,b}=\bm{J}+\bm{E}=(\bm{I}+\bm{N})\bm{J}$, its inverse is $\bm{J}^{-1}(\bm{I}-\bm{N})(\bm{I}+\bm{N}^2)$, recovering Eq.~\ref{eq:tinv-block}. Stacking $\bm{R}_b$ over $\bm{N}$ forms $\bm{Z}=\bm{R}_b\bm{N}$ and $\bm{N}^2$ in one product, and the last product multiplies $\bm{R}_b-\bm{Z}=\bm{R}_b(\bm{I}-\bm{N})$ by $\bm{I}+\bm{N}^2$. With two merge levels, the inverse takes eight products in total. The pseudocode specifies the algebraic dependencies; head packing, masked-product implementation, and instruction scheduling are omitted. Matrix products use BF16 operands with FP32 accumulation, so the returned matrix is an approximate inverse in floating-point arithmetic; \S\ref{sec:eval-e2e} evaluates application accuracy.

\begin{algorithm}[htbp]
\caption{\textsc{Inverse4}: hierarchical triangular inversion}
\label{alg:inverse4}
\small
\begin{algorithmic}[1]
\REQUIRE Unit lower-triangular $\bm{U}\in\mathbb{R}^{64\times64}$
\ENSURE Approximate inverse $\bm{R}\approx \bm{U}^{-1}$
\STATE $\bm{R}\leftarrow0$
\FOR{each consecutive index block $b$ of size $4$}
    \STATE $\bm{L}\leftarrow \bm{U}_{b,b}-\bm{I}$
    \STATE $\bm{W}\leftarrow \bm{L} \bm{L}$
    \STATE $\bm{R}_{b,b}\leftarrow(\bm{I}-\bm{L})(\bm{I}+\bm{W})$
\ENDFOR
\FOR{$h\in\{16,64\}$}
    \FOR{each consecutive index block $b$ of size $h$}
        \STATE Split $b$ into four equal subblocks $b_1,\ldots,b_4$
        \STATE $\bm{J}\leftarrow\operatorname{blockdiag}(\bm{U}_{b_1,b_1},\ldots,\bm{U}_{b_4,b_4})$
        \STATE $\bm{R}_b\leftarrow\operatorname{blockdiag}(\bm{R}_{b_1,b_1},\ldots,\bm{R}_{b_4,b_4})$
        \STATE $\bm{E}\leftarrow \bm{U}_{b,b}-\bm{J}$
        \STATE $\bm{N}\leftarrow \bm{E} \bm{R}_b$
        \STATE $[\bm{Z};\,\bm{N}^2]\leftarrow[\bm{R}_b;\,\bm{N}]\,\bm{N}$
        \STATE $\bm{R}_{b,b}\leftarrow (\bm{R}_b-\bm{Z})(\bm{I}+\bm{N}^2)$
    \ENDFOR
\ENDFOR
\STATE \textbf{return} $\bm{R}$
\end{algorithmic}
\end{algorithm}

Convolution presents a similar opportunity to replace vector work with matrix computation. In Figure~\ref{fig:vector}b, each tap contributes a scaled copy of the input, and a constant selector gathers the appropriate token rows and sums their contributions:
\begin{equation}
\bm{Y}=\bm{S}_{\mathrm{conv}}\bm{W}.
\label{eq:conv}
\end{equation}
For $\bm{X}\in\mathbb{R}^{C\times128}$ and $k$ taps, $\bm{W}\in\mathbb{R}^{kC\times128}$ contains the tap-scaled inputs and preceding-chunk halo, while $\bm{S}_{\mathrm{conv}}\in\{0,1\}^{C\times kC}$ performs selection and summation. This trades additional MXU arithmetic for fewer vector shifts and accumulations. The figure's selector $\bm{S}$ is distinct from the recurrent state.

\textbf{Exposing overlap across chunks.}
Matrix-based inversion removes dependent vector updates but leaves dependent matrix products. Executing chunks back to back would expose the latency gaps shown in the upper schedule of Figure~\ref{fig:schedule}b. The key opportunity is that the inverse depends only on the current chunk's inputs, not its incoming recurrent state. FlyWheel therefore overlaps the first chunk's inverse with the second chunk's stacked $\bm{A}_{kk}/\bm{A}_{qk}$ product, then overlaps the second inverse with the first chunk's state-dependent computation. Splitting this work across head groups fills gaps between inverse steps, as shown in the lower schedule, while preserving the order of recurrent state updates.

Algorithm~\ref{alg:FlyWheel-gdn} presents the computation in dependency order, with chunk index $c$ added to the notation of Eq.~\ref{eq:chunk}. Inputs follow convolution, SiLU, and query/key normalization; decay is omitted as in the Background. The unmasked products $\bm{R}_{kk,c},\bm{R}_{qk,c}\in\mathbb{R}^{C\times C}$ precede triangular masking; $\Delta\bm{S}_c\in\mathbb{R}^{d_k\times d_v}$ and $\bm{O}_{\mathrm{intra},c}\in\mathbb{R}^{C\times d_v}$ denote the state increment and intra-chunk output. Both value updates and outputs use the incoming $\bm{S}$ before the state update. Figure~\ref{fig:schedule}b shows how independent operations from adjacent chunks are interleaved.


\section{Evaluation}
\label{sec:eval}

We evaluate FlyWheel on TPU v6e and v7x, comparing softmax and GDN prefill kernels against TPU baselines and measuring application performance and accuracy on RULER, BABILong, and Video-MME. We isolate the contributions of individual designs through ablations in \S\ref{sec:discussion}.

\subsection{Setup}
\label{sec:eval-setup}

\textbf{Hardware and software.}
We run the kernel benchmarks on one TPU v6e (Trillium) chip and on one TensorCore of a TPU v7x (Ironwood) chip, which JAX exposes as one device (Table~\ref{tab:hardware}). The end-to-end runs use a v6e-8 slice with 8-way tensor parallelism (TP) up to 256K tokens, and a v6e-16 slice with TP 8 and 2-way prefill context parallelism (CP) at 512K tokens. On v7x they use one host of four chips, eight TensorCores, with TP 8 at every length, so each rank runs on one TensorCore. All runs use JAX~\cite{jax2018github} 0.11.0 and libtpu 0.0.44.

\begin{table}[!ht]
\caption{Hardware of the kernel benchmarks. A v7x chip has two TensorCores; the row lists one of them.}
\label{tab:hardware}
\centering
\small
\begin{tabular}{@{}lrrrr@{}}
\toprule
 & BF16 peak & VMEM & HBM & HBM BW \\
 & (TFLOP/s) & (MiB) & (GiB) & (TB/s) \\
\midrule
v6e chip       & 918  & 128 & 32 & 1.6 \\
v7x TensorCore & 1153 & 64  & 96 & 3.7 \\
\bottomrule
\end{tabular}
\end{table}

\textbf{Baselines.}
For softmax attention, we compare with \rpa{} (Ragged Paged Attention v3) from \tpuinf{}~\cite{tpuinference}, the TPU backend of vLLM~\cite{kwon2023efficient} (commit \texttt{4420cae}), and Google's \splash{} from \tokamax{}~\cite{tokamax} (commit \texttt{84e5f36}). For linear attention, we use \gdnv{}, the fused causal conv1d and chunked GDN kernel from the same \tpuinf{} commit.

\textbf{Kernel workloads.}
Softmax attention uses causal prefill with head dimension 256, 32 query heads, and 32 (MHA) or 4 (GQA) KV heads, for 1K to 128K tokens in powers of two. FlyWheel and \splash{} use $[H, T, D]$ inputs, while \rpa{} uses $[T, H, D]$ and a paged KV cache (following their original setup). To account for layout overhead, we measure both the kernel alone and the full attention block, including QKV and output projections and all intervening layout conversions (Appendix~\ref{app:softmax-block}).

Linear attention runs the fused causal conv1d (kernel size 4) and GDN prefill with 16 query/key heads, $d_k = d_v = 128$, and $n_v \in \{16, 32, 48\}$ value heads; Qwen3.8-4B has $n_v = 32$ and Qwen3.8-27B has $n_v = 48$. Each call prefills $T \in \{1\mathrm{K}, 4\mathrm{K}, 8\mathrm{K}, 16\mathrm{K}\}$ tokens, packed into 1 or 8 sequences of equal length. Activations are BF16 and the recurrent state is FP32. \gdnv{} runs its default 64-token tile; FlyWheel runs 128-row tiles of two 64-token chunks.

\textbf{End-to-end workloads.}
We serve Qwen3.8-27B~\cite{qwen38}, a hybrid vision-language model with 64 language model layers: 48 GDN layers (16 query/key heads, 48 value heads, head dimension 128) and 16 softmax attention layers (24 query heads, 4 KV heads, head dimension 256). We use vLLM (commit \texttt{51da0ca}) with \tpuinf{} (commit \texttt{4420cae}). FlyWheel replaces the baseline's \rpa{}, \gdnv{}, and vision encoder Pallas flash attention kernels. An environment switch selects the kernels within the same build; the model, scheduler, and other operators remain unchanged.

We also evaluate with three long-context benchmarks. RULER~\cite{hsieh2024ruler}, through lm-evaluation-harness~\cite{eval-harness}, runs its needle-in-a-haystack tasks at 4K, 32K, 128K, 256K, and 512K tokens. BABILong~\cite{kuratov2024babilong} runs tasks qa1--qa5 at 128K, 256K, and 512K tokens. Video-MME~\cite{fu2025video}, through lmms-eval~\cite{zhang2025lmms}, runs the 674 reasoning and synopsis questions on the long videos at 1024 and 2048 frames, about 113K and 225K visual tokens. At 512K tokens, and on v7x also at 256K, we extend RoPE with YaRN~\cite{peng2024yarn} at factor 2.

\textbf{Metrics.}
Kernel benchmarks report the median time per call over 5 rounds of back-to-back calls. Softmax kernel throughput is $4 \cdot B \cdot T^2 \cdot H \cdot D$ FLOPs, halved for causal masks, divided by device time from a profiler trace. The roofline uses an idealized traffic model that reads $\bm{Q}$, $\bm{K}$, and $\bm{V}$ once and writes $\bm{O}$ once, giving operational intensity $HT / (2(H + H_{kv}))$ FLOP/B for $H$ query heads and $H_{kv}$ KV heads. We measure full attention blocks in wall-clock time. GDN throughput is tokens per second for the complete call, including the wrapper's XLA operations.

For RULER and BABILong we report prefill throughput, the number of prompt tokens divided by the wall-clock time of generation. For Video-MME we report the mean time of the language model prefill per question. For accuracy we report each benchmark's own score.

\subsection{Softmax Attention}
\label{sec:eval-softmax}

\begin{figure}[t]
\centering
\includegraphics[width=\linewidth]{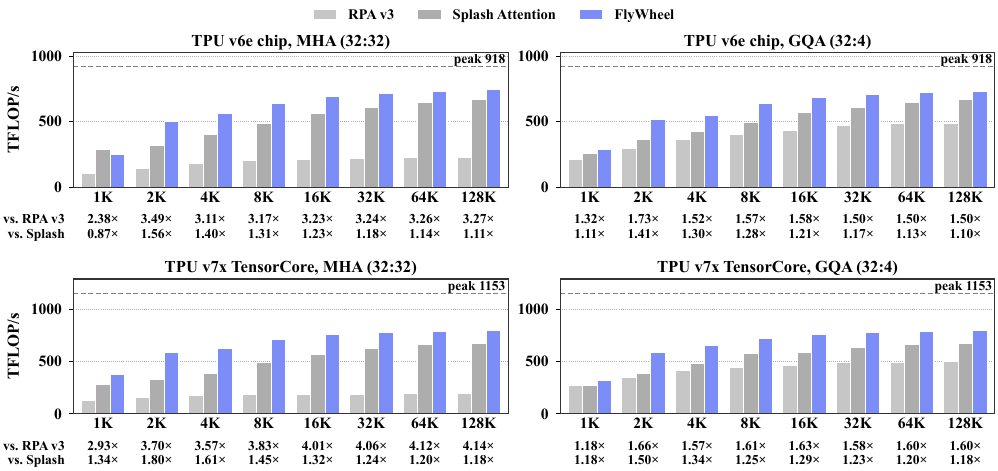}
\caption{Kernel-only throughput of softmax attention, causal prefill with head dimension 256, $B = 1$, and 32 query heads. Rows are one v6e chip and one v7x TensorCore; columns are MHA (32:32) and GQA (32:4). The numbers under each length are the speedups of FlyWheel over each baseline.}
\label{fig:softmax-kernel}
\end{figure}

\begin{figure}[t]
\centering
\includegraphics[width=\linewidth]{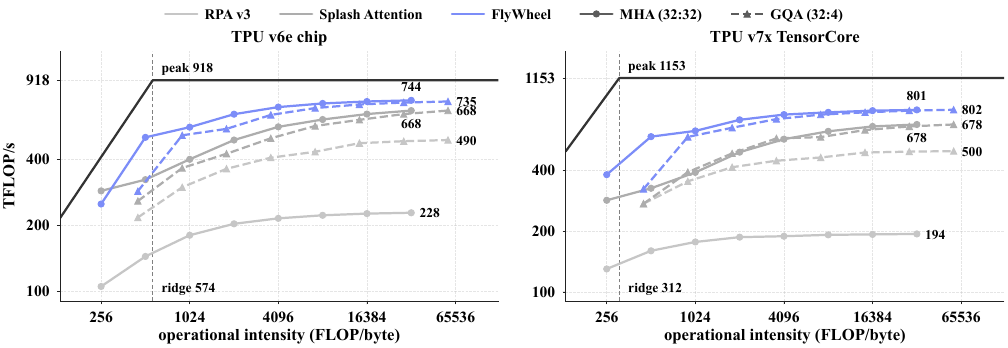}
\caption{The throughput of Figure~\ref{fig:softmax-kernel} on the roofline of one TPU v6e chip and one v7x TensorCore, for 1K to 128K tokens in powers of two. Solid lines with circles are MHA (32:32) and dashed lines with triangles are GQA (32:4). The number at the end of each line is its throughput at 128K tokens.}
\label{fig:roofline}
\end{figure}

Figure~\ref{fig:softmax-kernel} shows the throughput and FlyWheel's speedup at every shape. From 2K to 128K tokens, FlyWheel is 3.11--3.49$\times$ faster than \rpa{} under MHA and 1.50--1.73$\times$ under GQA on v6e, and 3.57--4.14$\times$ and 1.57--1.66$\times$ on v7x. Over the same range it is 1.10--1.56$\times$ faster than \splash{} on v6e and 1.18--1.80$\times$ on v7x.

\textbf{Roofline.}
Figure~\ref{fig:roofline} compares throughput against the roofline under the idealized traffic model of \S\ref{sec:eval-setup}. The ridge points are 574~FLOP/B on v6e and 312~FLOP/B on v7x. MHA at 1K and 2K and GQA at 1K on v6e, and MHA at 1K on v7x, lie on the bandwidth-limited branch; for example, the roof is 410~TFLOP/s for MHA at 1K on v6e. FlyWheel reaches up to 81\% of the roofline on v6e and 70\% on v7x (744 and 802~TFLOP/s), compared with 73\% and 59\% for \splash{} and 53\% and 43\% for \rpa{}. From 4K onward, all tested shapes lie on the compute-limited branch of this idealized model.

\textbf{Independence from the KV head count.}
From 8K onward, FlyWheel's MHA and GQA throughputs differ by at most 1.2\%, whereas \rpa{}'s MHA throughput is less than half its GQA throughput. This is consistent with the layout constraints in \S\ref{sec:design-softmax}: \rpa{} pads the query-head group to a multiple of two, wasting half the rows under MHA ($G = 1$), and its all-head tiles shrink as the head count grows. FlyWheel's per-head token tiles avoid both constraints.

\splash{} also maps tokens to sublanes and reaches 668~TFLOP/s under both MHA and GQA at 128K on v6e. FlyWheel further improves throughput through computation and scheduling changes; \S\ref{sec:disc-softmax} quantifies the contributions of the static anchor and fragment pipeline.

\textbf{Attention block.}
The gains extend to the full attention block, including projections and layout conversions: under MHA on v6e, FlyWheel is 1.48$\times$ faster than \rpa{} at 1K and 2.95$\times$ at 128K (Appendix~\ref{app:softmax-block}).

\textbf{Other shapes.}
A sweep over batch sizes, head counts, and head dimensions (Appendix~\ref{app:grid}) shows FlyWheel faster than both baselines in 785 of 792 cells, with throughput set by the total token count rather than the batch size.

\subsection{Linear Attention}
\label{sec:eval-linear}

\begin{figure}[t]
\centering
\includegraphics[width=\linewidth]{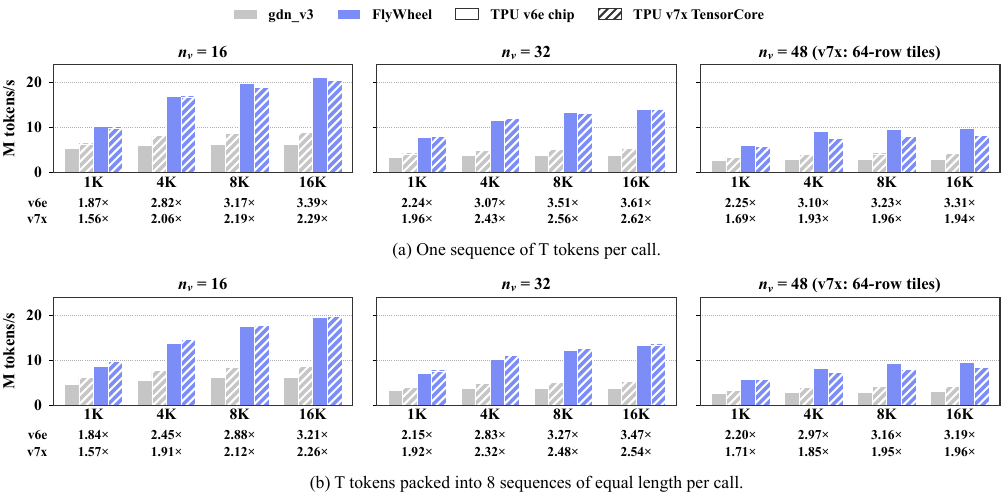}
\caption{Throughput of the fused causal conv1d and GDN prefill call, with 16 query/key heads and $d_k = d_v = 128$. Numbers under each length are FlyWheel's speedups over \gdnv{} on the same device. On v7x at $n_v = 48$, 64-row tiles fit in VMEM but can accommodate only a single chunk, preventing cross-chunk pipelining and resulting in a slight performance drop.}
\label{fig:gdn}
\end{figure}

Figure~\ref{fig:gdn} shows the GDN prefill throughput. On v6e, FlyWheel is 1.84--2.25$\times$ faster than \gdnv{} at 1K tokens, 2.45--3.10$\times$ at 4K, and 2.88--3.61$\times$ from 8K on. At $n_v = 48$, the shape of Qwen3.8-27B, it prefills 9.9M tokens/s at 16K against 3.0M for \gdnv{}. On one v7x TensorCore it is 1.56--2.62$\times$ faster.

\textbf{Layout work outside the kernel.}
\gdnv{} requires FP32 inputs in $[T, 1, \mathit{dim}]$ layout, adding wrapper operations for upcasting, retiling, and output reshaping. FlyWheel directly uses the projection layouts (\S\ref{sec:design-gdn}). At 8K on v6e, this change provides 1.52--1.68$\times$ speedup, with a further 2.04--2.10$\times$ from the three in-kernel designs (Table~\ref{tab:gdn-ablation}; \S\ref{sec:disc-gdn}).

\textbf{The speedup grows with the sequence length.}
Both kernels scale approximately linearly with $T$. On v6e at $n_v = 16$ with one sequence, \gdnv{} costs 163~$\mu$s per 1K tokens versus 45~$\mu$s for FlyWheel; across $n_v$, FlyWheel's per-token cost is 3.4--3.8$\times$ lower. Fixed overhead accounts for 7\% of FlyWheel's time at 16K, explaining why speedup increases with sequence length. Packing 8 sequences adds tens of microseconds to both kernels, with FlyWheel retaining a 1.84--3.47$\times$ speedup on v6e (Figure~\ref{fig:gdn}b).

\textbf{Comparison across TPU generations.}
For one sequence of 16K tokens, FlyWheel on a v6e chip is 2.3--2.6$\times$ faster than \gdnv{} on a v7x TensorCore across $n_v$, despite the latter's 2.3$\times$ HBM bandwidth and 1.26$\times$ peak FLOP/s (Table~\ref{tab:hardware}). At 16K and $n_v = 16$, FlyWheel uses 0.35~TB/s on v6e, about a fifth of its bandwidth. These results complement the ablations of layout and in-kernel overheads (Table~\ref{tab:gdn-ablation}). At $n_v = 48$, FlyWheel uses one 64-token chunk per tile to fit the 64~MiB VMEM of a v7x TensorCore and remains 1.69--1.96$\times$ faster than \gdnv{} on the same device.

\begin{figure}[t]
\centering
\includegraphics[width=\linewidth]{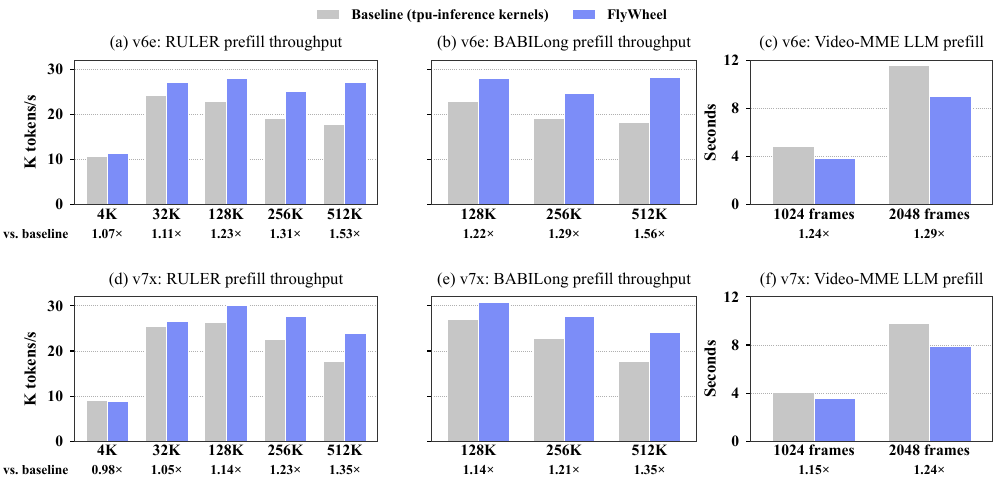}
\caption{Qwen3.8-27B in vLLM on v6e (top) and v7x (bottom). Numbers under each group are FlyWheel's speedups over the baseline. The v7x runs use 20 samples per RULER and BABILong task and 8 Video-MME questions.}
\label{fig:e2e}
\end{figure}

\subsection{End-to-End Application}
\label{sec:eval-e2e}

Figure~\ref{fig:e2e} shows that the kernel improvements translate into application-level gains on both TPU generations, with larger speedups at longer contexts.

\textbf{Text workloads on v6e.}
On RULER, FlyWheel's speedup increases from 1.07$\times$ at 4K to 1.23$\times$ at 128K and 1.53$\times$ at 512K. BABILong improves by 1.22--1.56$\times$ from 128K to 512K. This trend is consistent with softmax attention's growing share of prefill computation. From 32K to 256K on the v6e-8, baseline throughput falls by 21\% (24.3K to 19.1K tokens/s), while FlyWheel retains 93\% of its throughput (27.0K to 25.1K).

\textbf{Video-MME.}
On v6e, FlyWheel reduces language model prefill time from 4.80 to 3.87~s at 1024 frames (1.24$\times$) and from 11.60 to 9.00~s at 2048 frames (1.29$\times$; Figure~\ref{fig:e2e}c). These prompts contain about 113K and 225K visual tokens, respectively; the gains are comparable to RULER's 1.23$\times$ and 1.31$\times$ speedups at 128K and 256K (Figure~\ref{fig:e2e}a).

\textbf{Accuracy.}
On v6e, both runs achieve 100\% accuracy on all evaluated RULER tasks and lengths and identical BABILong scores. On Video-MME, both score 70.62 at 2048 frames; at 1024 frames, they differ by one of the 674 questions (71.66 versus 71.51).

\textbf{TPU v7x.}
On v7x, RULER speedup increases from 0.98$\times$ at 4K to 1.14$\times$ at 128K, 1.23$\times$ at 256K, and 1.35$\times$ at 512K, where throughput is 23.96K versus 17.70K tokens/s. BABILong improves by 1.14$\times$, 1.21$\times$, and 1.35$\times$ at the same lengths. Video-MME prefill improves by 1.15$\times$ at 1024 frames and 1.24$\times$ at 2048 (4.07 to 3.53~s and 9.77 to 7.88~s). On the evaluated samples, both runs give identical Video-MME answers, mean RULER scores differ by at most 0.8 points, and BABILong scores match at 128K and 256K and differ by one point at 512K.


\section{Ablation Study}
\label{sec:discussion}

We examine how FlyWheel's designs address the three bottlenecks identified in \S\ref{sec:bg-diff}: data layout and matrix operand reuse, vector work, and execution dependencies. The softmax ablations isolate computation and scheduling, while the GDN ablations evaluate all three aspects.

\subsection{Softmax Attention}
\label{sec:disc-softmax}

\begin{table}[t]
\caption{Softmax attention ablations on v6e: causal prefill at 32K tokens, $D = 256$, and 32 query and KV heads. Time is kernel device time. All configurations use the same data layout; the ablations vary softmax computation and fragment scheduling.}
\label{tab:softmax-ablation}
\centering
\small
\begin{tabular}{@{}lrr@{}}
\toprule
Configuration & Time (ms) & TFLOP/s \\
\midrule
Static anchor, pipeline (FlyWheel) & 24.43 & 720 \\
Online softmax, pipeline           & 30.21 & 582 \\
FA4 rescale, pipeline              & 30.86 & 570 \\
\midrule
Static anchor, no pipeline         & 33.79 & 521 \\
Online softmax, no pipeline        & 49.79 & 353 \\
\quad with the LP LLO scheduler    & 54.96 & 320 \\
FA4 rescale, no pipeline           & 50.54 & 348 \\
\bottomrule
\end{tabular}
\end{table}

Table~\ref{tab:softmax-ablation} compares the static-anchor softmax and fragment pipeline from \S\ref{sec:design-softmax}. We replace the static anchor with either online softmax or FlashAttention-4's conditional rescaling~\cite{zadouri2026flashattention}, which updates the running maximum only when the increase would scale the weights by more than $2^8$. Without pipelining, each fragment executes $\bm{Q}\bm{K}^{\top}$, softmax, and $\bm{P}\bm{V}$ consecutively.

\textbf{Reducing vector work.}
With pipelining enabled, replacing online softmax with the static anchor increases throughput from 582 to 720~TFLOP/s. Conditional rescaling reaches 570~TFLOP/s with the pipeline and 348~TFLOP/s without it, compared with 582 and 353~TFLOP/s for online softmax. It retains the per-fragment maximum reduction and rescale-factor computation, while its branch adds register pressure. The static anchor removes these recurring operations and is 26\% and 50\% faster than conditional rescaling with and without pipelining, respectively.

\textbf{Exposing independent work.}
The fragment pipeline increases static-anchor throughput from 521 to 720~TFLOP/s and online-softmax throughput from 353 to 582~TFLOP/s. Enabling Mosaic's LP LLO scheduler, used by \splash{} by default, does not recover this overlap in the unpipelined online-softmax implementation: throughput decreases to 320~TFLOP/s. This result supports explicitly interleaving independent fragments rather than relying on compiler scheduling of the original dependent loop.

Together, reduced vector work and fragment pipelining increase throughput from 353 to 720~TFLOP/s. Both are necessary to reach the full kernel's performance, which exceeds \splash{}'s 608~TFLOP/s by 18\% on this shape (Figure~\ref{fig:softmax-kernel}).

\subsection{Linear Attention}
\label{sec:disc-gdn}

Table~\ref{tab:gdn-ablation} incrementally adds FlyWheel's designs to \gdnv{}, yielding a total speedup of 3.19--3.53$\times$. These measurements include wrapper operations and report each design's additional benefit in the listed order, rather than its standalone contribution.

\textbf{Coordinating layouts and reusing operands.}
Coordinated layouts provide 1.52--1.68$\times$ speedup by eliminating the wrapper's HBM upcasting and retiling and producing output directly in the projection layout. Matrix operand reuse adds 1.06--1.22$\times$ by reducing repeated stationary-operand loads. These changes address both the surrounding data preparation and the efficiency of matrix execution.

\begin{table}[t]
\caption{Incremental GDN ablations on v6e: fused causal conv1d and GDN prefill of 8K tokens in one sequence, with 16 query/key heads and $d_k = d_v = 128$. Each design row reports speedup over the preceding configuration; their product gives the total speedup, up to rounding. All designs are described in \S\ref{sec:design-gdn}.}
\label{tab:gdn-ablation}
\centering
\small
\begin{tabular}{@{}lrrr@{}}
\toprule
 & \multicolumn{3}{c}{$n_v$} \\
\cmidrule(l){2-4}
Configuration / added design & 16 & 32 & 48 \\
\midrule
\gdnv{} time (ms)         & 1.31  & 2.15  & 2.75 \\
\midrule
Coordinated layouts      & 1.52  & 1.68  & 1.59 \\
Matrix operand reuse     & 1.06  & 1.22  & 1.16 \\
Reduced vector work      & 1.58  & 1.51  & 1.56 \\
Explicit pipelining      & 1.25  & 1.14  & 1.13 \\
\midrule
Total speedup            & 3.19  & 3.53  & 3.25 \\
FlyWheel time (ms)        & 0.412 & 0.610 & 0.847 \\
\bottomrule
\end{tabular}
\end{table}

\textbf{Reducing vector work.}
Reformulating the triangular inverse and convolution as MXU computations adds 1.51--1.58$\times$ speedup, together with implementations of SiLU and normalization that each use one EUP operation. Along with coordinated layouts, this is one of the two largest incremental gains, exceeding 1.5$\times$ at every tested value-head count.

\textbf{Overlapping independent computation.}
Explicit pipelining adds 1.13--1.25$\times$, with the largest benefit at $n_v = 16$. Packing two heads per matrix operand load produces $n_v / 2$ independent inverse chains: 8 chains at $n_v = 16$, and 16 and 24 chains at $n_v = 32$ and 48. With fewer chains, more inverse latency remains exposed, so pipelining gains 25\%, while operand reuse gains 6\%. At larger head counts, the additional chains provide more independent work: pipelining gains 13--14\%, while operand reuse gains 16--22\%. The two incremental factors together contribute 1.31--1.39$\times$, showing how operand reuse and scheduling address different limits as the available parallelism changes.


\section{Related Work}
\label{sec:related}

\textbf{Attention kernels on GPUs and other accelerators.}
FlashAttention~\cite{dao2022flashattention} computes attention in one tiled kernel with online softmax~\cite{milakov2018online}. Its successors split the work across warps~\cite{dao2024flashattention}, overlap the softmax of one warpgroup with the matmuls of another~\cite{shah2024flashattention}, and co-design the softmax and the pipeline for Blackwell~\cite{zadouri2026flashattention}. Flash-Decoding~\cite{dao2023flashdecoding} and FlashDecoding++~\cite{hong2024flashdecoding++} split the KV sequence of a decode step, and FlashInfer~\cite{ye2025flashinfer} stores paged and ragged KV caches as block-sparse matrices. FuseMax~\cite{nayak2024fusemax} and MAS-Attention~\cite{shakerdargah2025mas} overlap the matmuls and the softmax on a spatial accelerator and an edge NPU, whose matrix and vector engines run concurrently, as the warpgroups of a GPU do. A TensorCore issues both from one instruction stream (\S\ref{sec:bg-diff}), so FlyWheel writes the overlap into the program order (\S\ref{sec:design}).

\textbf{Attention kernels on TPU.}
\rpa{}~\cite{jiang2026ragged}, the softmax attention kernel of \tpuinf{}, serves prefill, decode, and mixed ragged batches. It puts heads on sublanes so that a sequence can start at any token, reads K and V from a paged cache that interleaves their heads, fuses the KV cache update into its DMA pipeline, and overlaps the $\bm{Q}\bm{K}^\top$ and softmax of one KV head with the $\bm{P}\bm{V}$ of the previous one; \S\ref{sec:design-softmax} discusses what this layout and this pipeline cost. \splash{}~\cite{tokamax} puts tokens on sublanes like FlyWheel and runs online softmax over the K/V fragments; FlyWheel's gain over it comes from that loop (\S\ref{sec:disc-softmax}).

\textbf{Softmax without the rescale.}
Several kernels replace the running maximum with a fixed value. FlashDecoding++~\cite{hong2024flashdecoding++} takes one value per model from offline statistics of the scores, checks each score against a window around it, and recomputes the row with synchronized softmax when a score falls outside. VFA~\cite{sun2026vfa} sets each row's maximum from key-block summaries and the exact maxima of the first and the local block, then freezes it without a check. \splash{} accepts a constant maximum from the caller, also without a check. Concurrently with this work, Stashuk et al.~\cite{stashuk2026efficient} anchor each row of a causal training kernel on Blackwell at the maximum of its first key block, check each 128-row group, and recompute the failing tiles in a second launch. FlyWheel anchors each row on a score the row attains, so it needs no profile, key summary, or value from the caller. Its check tests no score: the FP32 sum that the kernel already keeps bounds every weight of the row, so one comparison per row certifies the row (\S\ref{sec:design-softmax}), and a failing block is replayed within the same launch.

\textbf{Chunked kernels for linear attention.}
Chunked prefill kernels for DeltaNet and GDN~\cite{yang2024parallelizing,yang2025gated} differ in how they invert the unit lower-triangular matrix of each chunk. FLA~\cite{yang2024fla} inverts $16 \times 16$ diagonal blocks by forward substitution and merges them with tensor-core matmuls, and \gdnv{} runs forward substitution on the VPU. Sobczyk et al.~\cite{sobczyk2026fast} invert with matmuls alone: a truncated Neumann series on diagonal blocks of 16 or 32 rows, merged by radix-2 doubling with FP32 products, with a proof of stability and kernels for Ascend NPUs. The Kimi Delta Attention (KDA)~\cite{team2025kimi} kernel of \tokamax{} applies a Neumann series to $8 \times 8$ diagonal blocks and again to the block-nilpotent remainder, on the TPU MXU with FP32 operands. FlyWheel's inverse is a radix-4 member of this family that replaces FP32-operand products with BF16 matmuls. Starting from $4 \times 4$ diagonal blocks, it avoids forming high powers of the full matrix and computes the inverse of a 64-token chunk using eight BF16 matmuls with FP32 accumulation. This design maintains comparable accuracy to the baseline on RULER and BABILong with Qwen3.8-27B (\S\ref{sec:eval-e2e}).


\section{Conclusion}
\label{sec:conclusion}

We present FlyWheel, TPU-native kernels for softmax and linear attention. FlyWheel combines operand reuse, reduced vector work, and explicit pipelining, with a checked static-anchor softmax and a BF16 triangular inverse as key computational designs. Evaluations on TPU v6e and v7x demonstrate kernel and application-level speedups while maintaining comparable model accuracy.

\section*{Acknowledgments}
This work was supported by a 2026 Google TPU Research Award and the TPU Builders Program. The authors thank Josh Gordon, Lauren Jensen, Edgar Chen, Satha Kumar, Tasha Allen, Aditi Joshi, and the entire TPU Builders Support Team for their help with TPU onboarding.

\bibliographystyle{plainnat}
\bibliography{references}

@misc{jax2018github,
  author = {James Bradbury and Roy Frostig and Peter Hawkins and Matthew James Johnson and Yash Katariya and Chris Leary and Dougal Maclaurin and George Necula and Adam Paszke and Jake Vander{P}las and Skye Wanderman-{M}ilne and Qiao Zhang},
  title = {{JAX}: composable transformations of {P}ython+{N}um{P}y programs},
  url = {http://github.com/jax-ml/jax},
  version = {0.3.13},
  year = {2018},
}

@inproceedings{kwon2023efficient,
  title={Efficient memory management for large language model serving with pagedattention},
  author={Kwon, Woosuk and Li, Zhuohan and Zhuang, Siyuan and Sheng, Ying and Zheng, Lianmin and Yu, Cody Hao and Gonzalez, Joseph and Zhang, Hao and Stoica, Ion},
  booktitle={Proceedings of the 29th symposium on operating systems principles},
  pages={611--626},
  year={2023}
}

@misc{tpuinference,
  author       = {{vLLM Project}},
  title        = {tpu-inference: {TPU} backend for {vLLM}},
  year         = {2026},
  howpublished = {\url{https://github.com/vllm-project/tpu-inference}}
}

@misc{tokamax,
  author       = {{OpenXLA}},
  title        = {Tokamax: a library of custom accelerator kernels},
  year         = {2026},
  howpublished = {\url{https://github.com/openxla/tokamax}}
}

@misc{qwen38,
    title = {{Qwen3.8-Max}: A New Bar for Coding and Cowork},
    url = {https://qwen.ai/blog?id=qwen3.8},
    author = {{Qwen Team}},
    month = {August},
    year = {2026}
}

@article{hsieh2024ruler,
  title={RULER: What's the real context size of your long-context language models?},
  author={Hsieh, Cheng-Ping and Sun, Simeng and Kriman, Samuel and Acharya, Shantanu and Rekesh, Dima and Jia, Fei and Zhang, Yang and Ginsburg, Boris},
  journal={arXiv preprint arXiv:2404.06654},
  year={2024}
}

@article{kuratov2024babilong,
  title={Babilong: Testing the limits of llms with long context reasoning-in-a-haystack},
  author={Kuratov, Yuri and Bulatov, Aydar and Anokhin, Petr and Rodkin, Ivan and Sorokin, Dmitry and Sorokin, Artyom and Burtsev, Mikhail},
  journal={Advances in Neural Information Processing Systems},
  volume={37},
  pages={106519--106554},
  year={2024}
}

@inproceedings{fu2025video,
  title={Video-mme: The first-ever comprehensive evaluation benchmark of multi-modal llms in video analysis},
  author={Fu, Chaoyou and Dai, Yuhan and Luo, Yongdong and Li, Lei and Ren, Shuhuai and Zhang, Renrui and Wang, Zihan and Zhou, Chenyu and Shen, Yunhang and Zhang, Mengdan and others},
  booktitle={2025 IEEE/CVF Conference on Computer Vision and Pattern Recognition (CVPR)},
  pages={24108--24118},
  year={2025},
  organization={IEEE}
}

@misc{eval-harness,
  author       = {Gao, Leo and Tow, Jonathan and Abbasi, Baber and Biderman, Stella and Black, Sid and DiPofi, Anthony and Foster, Charles and Golding, Laurence and Hsu, Jeffrey and Le Noac'h, Alain and Li, Haonan and McDonell, Kyle and Muennighoff, Niklas and Ociepa, Chris and Phang, Jason and Reynolds, Laria and Schoelkopf, Hailey and Skowron, Aviya and Sutawika, Lintang and Tang, Eric and Thite, Anish and Wang, Ben and Wang, Kevin and Zou, Andy},
  title        = {The Language Model Evaluation Harness},
  month        = 07,
  year         = 2024,
  publisher    = {Zenodo},
  version      = {v0.4.3},
  doi          = {10.5281/zenodo.12608602},
  url          = {https://zenodo.org/records/12608602}
}

@inproceedings{zhang2025lmms,
  title={Lmms-eval: Reality check on the evaluation of large multimodal models},
  author={Zhang, Kaichen and Li, Bo and Zhang, Peiyuan and Pu, Fanyi and Cahyono, Joshua Adrian and Hu, Kairui and Liu, Shuai and Zhang, Yuanhan and Yang, Jingkang and Li, Chunyuan and others},
  booktitle={Findings of the Association for Computational Linguistics: NAACL 2025},
  pages={881--916},
  year={2025}
}

@inproceedings{peng2024yarn,
  title={Yarn: Efficient context window extension of large language models},
  author={Peng, Bowen and Quesnelle, Jeffrey and Fan, Honglu and Shippole, Enrico},
  booktitle={International Conference on Learning Representations},
  volume={2024},
  pages={31932--31951},
  year={2024}
}

@article{zadouri2026flashattention,
  title={Flashattention-4: Algorithm and kernel pipelining co-design for asymmetric hardware scaling},
  author={Zadouri, Ted and Hoehnerbach, Markus and Shah, Jay and Thakkar, Vijay and Dao, Tri},
  journal={Proceedings of Machine Learning and Systems},
  volume={8},
  pages={912--926},
  year={2026}
}

@article{sobczyk2026fast,
  title={Fast and Stable Triangular Inversion for Delta-Rule Linear Transformers},
  author={Sobczyk, Aleksandros and Gottardo, Gioele and Matzoros, Christos K and De Vita, Mirko and Skogh, Filip and Zouzias, Anastasios and Zhuang, Jiawei},
  journal={arXiv preprint arXiv:2605.21325},
  year={2026}
}

@article{shah2024flashattention,
  title={Flashattention-3: Fast and accurate attention with asynchrony and low-precision},
  author={Shah, Jay and Bikshandi, Ganesh and Zhang, Ying and Thakkar, Vijay and Ramani, Pradeep and Dao, Tri},
  journal={Advances in Neural Information Processing Systems},
  volume={37},
  pages={68658--68685},
  year={2024}
}

@article{team2025kimi,
  title={Kimi linear: An expressive, efficient attention architecture},
  author={Team, Kimi and Zhang, Yu and Lin, Zongyu and Yao, Xingcheng and Hu, Jiaxi and Meng, Fanqing and Liu, Chengyin and Men, Xin and Yang, Songlin and Li, Zhiyuan and others},
  journal={arXiv preprint arXiv:2510.26692},
  year={2025}
}

@article{vaswani2017attention,
  title={Attention is all you need},
  author={Vaswani, Ashish and Shazeer, Noam and Parmar, Niki and Uszkoreit, Jakob and Jones, Llion and Gomez, Aidan N and Kaiser, {\L}ukasz and Polosukhin, Illia},
  journal={Advances in neural information processing systems},
  volume={30},
  year={2017}
}

@inproceedings{ainslie2023gqa,
  title={Gqa: Training generalized multi-query transformer models from multi-head checkpoints},
  author={Ainslie, Joshua and Lee-Thorp, James and De Jong, Michiel and Zemlyanskiy, Yury and Lebr{\'o}n, Federico and Sanghai, Sumit},
  booktitle={Proceedings of the 2023 conference on empirical methods in natural language processing},
  pages={4895--4901},
  year={2023}
}

@article{dao2022flashattention,
  title={Flashattention: Fast and memory-efficient exact attention with io-awareness},
  author={Dao, Tri and Fu, Dan and Ermon, Stefano and Rudra, Atri and R{\'e}, Christopher},
  journal={Advances in neural information processing systems},
  volume={35},
  pages={16344--16359},
  year={2022}
}

@inproceedings{dao2024flashattention,
  title={Flashattention-2: Faster attention with better parallelism and work partitioning},
  author={Dao, Tri},
  booktitle={International Conference on Learning Representations},
  volume={2024},
  pages={35549--35562},
  year={2024}
}

@article{milakov2018online,
  title={Online normalizer calculation for softmax},
  author={Milakov, Maxim and Gimelshein, Natalia},
  journal={arXiv preprint arXiv:1805.02867},
  year={2018}
}

@article{hong2024flashdecoding++,
  title={Flashdecoding++: Faster large language model inference with asynchronization, flat gemm optimization, and heuristics},
  author={Hong, Ke and Dai, Guohao and Xu, Jiaming and Mao, Qiuli and Li, Xiuhong and Liu, Jun and Chen, Kangdi and Dong, Yuhan and Wang, Yu},
  journal={Proceedings of Machine Learning and Systems},
  volume={6},
  pages={148--161},
  year={2024}
}

@article{ye2025flashinfer,
  title={Flashinfer: Efficient and customizable attention engine for llm inference serving},
  author={Ye, Zihao and Chen, Lequn and Lai, Ruihang and Lin, Wuwei and Zhang, Yineng and Wang, Stephanie and Chen, Tianqi and Kasikci, Baris and Grover, Vinod and Krishnamurthy, Arvind and others},
  journal={Proceedings of Machine Learning and Systems},
  volume={7},
  year={2025}
}

@inproceedings{katharopoulos2020transformers,
  title={Transformers are rnns: Fast autoregressive transformers with linear attention},
  author={Katharopoulos, Angelos and Vyas, Apoorv and Pappas, Nikolaos and Fleuret, Fran{\c{c}}ois},
  booktitle={International conference on machine learning},
  pages={5156--5165},
  year={2020},
  organization={PMLR}
}

@inproceedings{schlag2021linear,
  title={Linear transformers are secretly fast weight programmers},
  author={Schlag, Imanol and Irie, Kazuki and Schmidhuber, J{\"u}rgen},
  booktitle={International conference on machine learning},
  pages={9355--9366},
  year={2021},
  organization={PMLR}
}

@article{yang2024parallelizing,
  title={Parallelizing linear transformers with the delta rule over sequence length},
  author={Yang, Songlin and Wang, Bailin and Zhang, Yu and Shen, Yikang and Kim, Yoon},
  journal={Advances in neural information processing systems},
  volume={37},
  pages={115491--115522},
  year={2024}
}

@inproceedings{yang2025gated,
  title={Gated delta networks: Improving mamba2 with delta rule},
  author={Yang, Songlin and Kautz, Jan and Hatamizadeh, Ali},
  booktitle={International Conference on Learning Representations},
  volume={2025},
  pages={29687--29707},
  year={2025}
}

@article{jiang2026ragged,
  title={Ragged paged attention: A high-performance and flexible llm inference kernel for tpu},
  author={Jiang, Jevin and Chen, Ying and Hechtman, Blake A and Zhang, Fenghui and Mu, Yarong},
  journal={arXiv preprint arXiv:2604.15464},
  year={2026}
}

@article{jouppi2020domain,
  title={A domain-specific supercomputer for training deep neural networks},
  author={Jouppi, Norman P and Yoon, Doe Hyun and Kurian, George and Li, Sheng and Patil, Nishant and Laudon, James and Young, Cliff and Patterson, David},
  journal={Communications of the ACM},
  volume={63},
  number={7},
  pages={67--78},
  year={2020},
  publisher={ACM, New York, NY, USA}
}

@article{jouppi2026google,
  title={Google's Training Supercomputers from TPUv2 to Ironwood: Architectural Stability, Scale, Resilience, Power Efficiency, and Sustainability Across Five Generations},
  author={Jouppi, Norman P and Lakshmanamurthy, Sridhar and Young, Cliff and Patterson, David},
  journal={IEEE Micro},
  year={2026},
  publisher={IEEE}
}

@misc{dao2023flashdecoding,
  author       = {Tri Dao and Daniel Haziza and Francisco Massa and Grigory Sizov},
  title        = {Flash-Decoding for Long-Context Inference},
  year         = {2023},
  howpublished = {\url{https://crfm.stanford.edu/2023/10/12/flashdecoding.html}}
}

@misc{yang2024fla,
  title  = {FLA: A Triton-Based Library for Hardware-Efficient Implementations of Linear Attention Mechanism},
  author = {Yang, Songlin and Zhang, Yu},
  url    = {https://github.com/fla-org/flash-linear-attention},
  month  = jan,
  year   = {2024}
}

@article{sun2026vfa,
  title={VFA: Relieving Vector Operations in Flash Attention with Global Maximum Pre-computation},
  author={Sun, Yupeng and Li, Yanzhao and Zou, Zhiqiang and Du, Bai and Zhang, Zhiyuan and Dong, Hui and Fan, Gaoyige and Wang, Hui},
  journal={arXiv preprint arXiv:2604.12798},
  year={2026}
}

@article{stashuk2026efficient,
  title={Efficient FlashAttention on Blackwell via Fixed-Shift Softmax and Persistent Scheduling},
  author={Stashuk, Oleksandr and Yu, Hongtao and Shah, Jay},
  journal={arXiv preprint arXiv:2610.02229},
  year={2026}
}

@article{shakerdargah2025mas,
  title={MAS-Attention: Memory-Aware Stream Processing for Attention Acceleration on Resource-Constrained Edge Devices},
  author={Shakerdargah, Mohammadali and Lu, Shan and Gao, Chao and Niu, Di},
  journal={Proceedings of Machine Learning and Systems},
  volume={7},
  year={2025}
}

@inproceedings{nayak2024fusemax,
  title={Fusemax: Leveraging extended einsums to optimize attention accelerator design},
  author={Nayak, Nandeeka and Wu, Xinrui and Odemuyiwa, Toluwanimi O and Pellauer, Michael and Emer, Joel S and Fletcher, Christopher W},
  booktitle={2024 57th IEEE/ACM International Symposium on Microarchitecture (MICRO)},
  pages={1458--1473},
  year={2024},
  organization={IEEE}
}

@misc{pallas,
  author       = {{The JAX Authors}},
  title        = {Pallas: a {JAX} kernel language},
  year         = {2026},
  howpublished = {\url{https://docs.jax.dev/en/latest/pallas/index.html}}
}

@misc{semianalysis2026inferencex,
  author       = {Alec Ibarra and Cam Quilici and Bryan Shan and others},
  title        = {{TPU} Inference Externalization Full Steam Ahead: {InferenceX}},
  year         = {2026},
  howpublished = {SemiAnalysis, \url{https://newsletter.semianalysis.com/p/tpu-inferencex-full-steam}}
}

@misc{liou2026replayssmcachessminputs,
      title={ReplaySSM: Cache SSM Inputs, Not State}, 
      author={Ze-Wei Liou and Harris Nover and Amey Naik and Andrii Skliar and Igor Shovkun and Ido Shahaf and Tri Dao},
      year={2026},
      eprint={2610.13778},
      archivePrefix={arXiv},
      primaryClass={cs.LG},
      url={https://arxiv.org/abs/2610.13778}, 
}

@misc{vllmtorchtpu,
  author       = {{vLLM Project}},
  title        = {{vLLM TorchTPU}},
  year         = {2026},
  howpublished = {\url{https://github.com/vllm-project/vllm-torchtpu}}
}

\newpage
\appendix

\section{Reference Attention Algorithms}
\label{app:reference-algorithms}

Algorithms~\ref{alg:fa-reference} and~\ref{alg:gdn-reference} provide mathematical reference formulations for the two attention computations in \S\ref{sec:background}. They expose the online rescaling and triangular substitution replaced by FlyWheel; they do not reproduce the full scheduling or implementation details of a particular library. Both algorithms show one head, omit the attention scaling and GDN decay factors as in the main text, and assume complete tiles for clarity. \textsc{Scores} includes causal masking, with at least one valid key per query row in the first key tile. Row-wise vectors are broadcast across columns where required.

\begin{algorithm}[htbp]
\caption{FlashAttention reference: blockwise online softmax}
\label{alg:fa-reference}
\small
\begin{algorithmic}[1]
\REQUIRE $\bm{Q},\bm{K},\bm{V}\in\mathbb{R}^{T\times D}$
\ENSURE $\bm{O}\in\mathbb{R}^{T\times D}$
\FOR{each query tile $\bm{Q}_i$}
    \STATE Partition keys and values into $(\bm{K}_j,\bm{V}_j)_{j=1}^{J}$
    \STATE $\bm{s}\leftarrow\textsc{Scores}(\bm{Q}_i,\bm{K}_1)$
    \STATE $\bm{m}\leftarrow\operatorname{rowmax}(\bm{s})$; $\bm{p}\leftarrow e^{\bm{s}-\bm{m}}$
    \STATE $\bm{l}\leftarrow\operatorname{rowsum}(\bm{p})$; $\bm{o}\leftarrow \bm{p}\bm{V}_1$
    \FOR{$j=2,\ldots,J$}
        \STATE $\bm{s}\leftarrow\textsc{Scores}(\bm{Q}_i,\bm{K}_j)$
        \STATE $\bm{m}'\leftarrow\max(\bm{m},\operatorname{rowmax}(\bm{s}))$
        \STATE $\bm{r}\leftarrow e^{\bm{m}-\bm{m}'}$; $\bm{p}\leftarrow e^{\bm{s}-\bm{m}'}$
        \STATE $\bm{l}\leftarrow \bm{r}\odot \bm{l}+\operatorname{rowsum}(\bm{p})$
        \STATE $\bm{o}\leftarrow \bm{r}\odot \bm{o}+\bm{p}\bm{V}_j$
        \STATE $\bm{m}\leftarrow \bm{m}'$
    \ENDFOR
    \STATE $\bm{O}_i\leftarrow \bm{o}/\bm{l}$
\ENDFOR
\STATE \textbf{return} $\bm{O}$
\end{algorithmic}
\end{algorithm}

\textbf{Relation to FlyWheel softmax.}
Algorithm~\ref{alg:fa-reference} follows the blockwise online-softmax computation used by FlashAttention~\cite{dao2022flashattention,milakov2018online}. For a query tile with $B_q$ rows and key tile with $B_k$ rows, $\bm{s},\bm{p}\in\mathbb{R}^{B_q\times B_k}$, $\bm{m},\bm{l},\bm{r}\in\mathbb{R}^{B_q}$, and $\bm{o}\in\mathbb{R}^{B_q\times D}$. Each fragment updates the maximum and rescales the previous accumulators. Algorithm~\ref{alg:FlyWheel-softmax} replaces these repeated updates with a checked fixed anchor and pipelines independent fragment computation; Algorithm~\ref{alg:guarded-replay} supplies its fallback.

\begin{algorithm}[htbp]
\caption{Chunked GDN reference with forward substitution}
\label{alg:gdn-reference}
\small
\begin{algorithmic}[1]
\REQUIRE $\bm{Q},\bm{K}\in\mathbb{R}^{T\times d_k}$, $\bm{V}\in\mathbb{R}^{T\times d_v}$; coefficients $\beta_t$; state $\bm{S}\in\mathbb{R}^{d_k\times d_v}$
\ENSURE $\bm{O}\in\mathbb{R}^{T\times d_v}$ and updated state $\bm{S}$
\STATE Partition inputs into chunks of $C$ tokens
\FOR{each chunk $c$}
    \STATE $\bm{\beta}_c\leftarrow\operatorname{diag}(\beta_{(c-1)C+1},\ldots,\beta_{cC})$
    \STATE $\bm{A}_{kk,c}\leftarrow\operatorname{tril}_{-1}(\bm{\beta}_c\bm{K}_c\bm{K}_c^\top)$
    \STATE $\bm{A}_{qk,c}\leftarrow\operatorname{tril}_{0}(\bm{Q}_c\bm{K}_c^\top)$
    \STATE $\bm{R}\leftarrow \bm{I}$
    \FOR{$t=1,\ldots,C$}
        \STATE $\bm{R}_{t,:}\leftarrow \bm{I}_{t,:}-\sum_{j=1}^{t-1}(\bm{A}_{kk,c})_{tj}\bm{R}_{j,:}$
    \ENDFOR
    \STATE $\bm{V}_{\mathrm{new}}\leftarrow \bm{R}\bm{\beta}_c(\bm{V}_c-\bm{K}_c\bm{S})$
    \STATE $\bm{O}_c\leftarrow \bm{Q}_c\bm{S}+\bm{A}_{qk,c}\bm{V}_{\mathrm{new}}$
    \STATE $\bm{S}\leftarrow \bm{S}+\bm{K}_c^\top \bm{V}_{\mathrm{new}}$
\ENDFOR
\STATE \textbf{return} $\bm{O},\bm{S}$
\end{algorithmic}
\end{algorithm}

\textbf{Relation to FlyWheel GDN.}
Algorithm~\ref{alg:gdn-reference} implements the chunked recurrence of Eq.~\ref{eq:chunk}~\cite{yang2024parallelizing,yang2025gated}, with inputs after convolution, SiLU, and query/key normalization. Forward substitution constructs $\bm{R}=(\bm{I}+\bm{A}_{kk,c})^{-1}\in\mathbb{R}^{C\times C}$ one row at a time; $(\bm{A}_{kk,c})_{tj}$ is a scalar matrix entry. Algorithm~\ref{alg:FlyWheel-gdn} instead combines products sharing an operand and uses the matrix-based inverse in Algorithm~\ref{alg:inverse4}. Its implementation additionally overlaps chunk-local work with state-dependent updates, as described in \S\ref{sec:design-gdn}.

\section{Softmax Attention Block Time}
\label{app:softmax-block}

\begin{figure}[htbp]
\centering
\includegraphics[width=\textwidth]{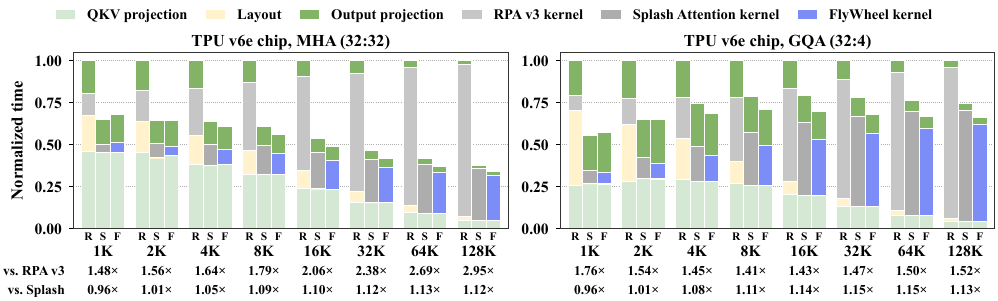}
\caption{Softmax attention block time on one v6e chip, from hidden states to
hidden states ($d_{\mathrm{model}} = 8192$), causal prefill with head
dimension 256, $B = 1$, and 32 query heads, split by segment and normalized
to the slowest block of each group. Panels are MHA (32:32) and GQA (32:4).
R, S, and F are \rpa{}, \splash{}, and FlyWheel; the numbers under each
length are the wall-clock speedups of FlyWheel's block over each baseline.}
\label{fig:softmax-block}
\end{figure}

Figure~\ref{fig:softmax-block} splits the time of the attention block on v6e (\S\ref{sec:eval-setup}) into the QKV projection, the layout work, the kernel, and the output projection. FlyWheel and \splash{} need no layout work: the projections write $[H, T, D]$ and the output projection reads it (\S\ref{sec:design-softmax}). \rpa{} reads K and V from a paged cache that interleaves heads, so its block concatenates and transposes them before the kernel, which costs 0.25~ms at 1K and 33.7~ms at 128K under MHA. FlyWheel's block is 1.48$\times$ faster than \rpa{}'s at 1K and 2.95$\times$ at 128K under MHA (466 against 1371~ms), and 1.41--1.76$\times$ under GQA. The speedup grows with the length because the projections are linear in $T$ and attention is quadratic: attention is 3\% of the block's FLOPs at 1K and 80\% at 128K, and FlyWheel's kernel takes 8\% of its MHA block time at 1K and 80\% at 128K. The layout work therefore sets 65\% of the MHA gap at 1K, and the kernel sets 96\% of it at 128K. On v7x the block speedups over \rpa{} are 1.50--3.62$\times$ under MHA and 1.29--1.55$\times$ under GQA. \splash{} does no layout work either, so FlyWheel's block speedup over it is smaller: 1.01--1.15$\times$ from 2K on. At 1K, where the projections take about 90\% of either block, \splash{}'s block is 4\% faster.

\section{The Full Workload Grid}
\label{app:grid}

The kernel measurements of \S\ref{sec:eval-softmax} fix $B = 1$, $D = 256$,
$H = 32$, and $H_{kv} \in \{32, 4\}$. To check that the comparison does not
depend on that choice, we sweep the Cartesian product of $H \in \{16, 32\}$,
$D \in \{128, 256\}$, every $H_{kv}$ from $H$ down to 1 in powers of two,
$B \in \{1, \dots, 128\}$, and $T \in \{1\mathrm{K}, \dots, 128\mathrm{K}\}$,
keeping the cells with $BT \le 128$K: 2376 measurements over 22 head
configurations $\times$ 36 (batch, length) cells $\times$ 3 implementations, on
one v6e chip, causal, BF16. Throughput follows \S\ref{sec:eval-setup} --- the
kernel figures divide the analytic FLOPs by the Pallas call's device time in a
profiler trace, the block figures divide the block's FLOPs, projections
included, by its wall clock.

\begin{figure}[htbp]
\centering
\includegraphics[width=\textwidth]{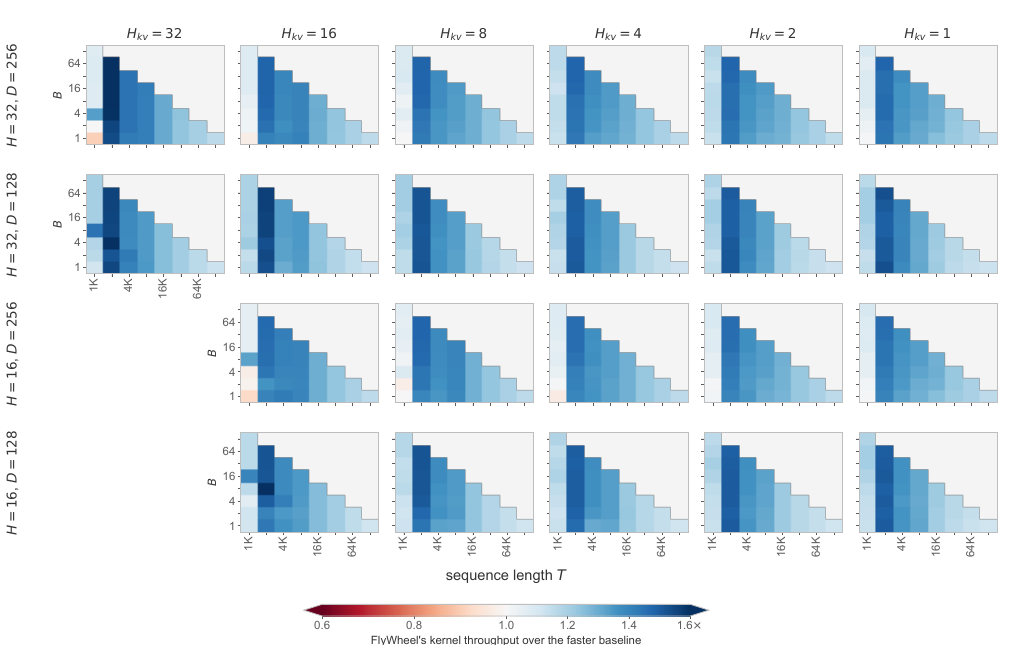}
\caption{FlyWheel's kernel throughput over the faster of \rpa{} and \splash{}
on one v6e chip, at every cell of the grid: 22 combinations of query heads
$H$, KV heads $H_{kv}$, and head dimension $D$, each swept over 8 batch sizes
and 8 sequence lengths with $BT \le 128$K. Blue is FlyWheel ahead, red behind.
Columns share a KV head count, so the $H_{kv} = 32$ slots are empty for
$H = 16$; the pale corner of each panel is the part of the rectangle the token
budget excludes.}
\label{fig:grid-speedup}
\end{figure}

\textbf{Across the grid.}
The advantage of \S\ref{sec:eval-softmax} does not depend on the shape it was
measured at. FlyWheel's kernel is faster than both baselines in 785 of the 792
cells, by a median of $1.30\times$, and in all 330 cells from 8K tokens on, by
a median of $1.24\times$ and never by less than $1.12\times$. The seven
exceptions are all at $T = 1$K with $D = 256$, where FlyWheel stays within
9\% of \splash{}. Neither the KV head count nor the batch size changes the
picture: in Figure~\ref{fig:grid-speedup} the panels are blue throughout.

\begin{figure}[htbp]
\centering
\includegraphics[width=\textwidth]{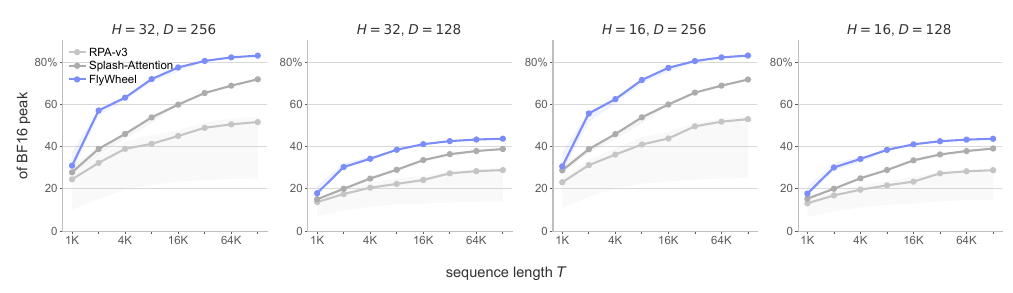}
\caption{Kernel utilization of the v6e BF16 peak over the same grid. Each line
is the median across all KV head counts and batch sizes at that length, and
the band is their full range. \rpa{}'s band is wide because its throughput
depends on the KV head count (\S\ref{sec:eval-softmax}); FlyWheel's and
\splash{}'s are narrow.}
\label{fig:grid-util}
\end{figure}

\textbf{Utilization.}
Figure~\ref{fig:grid-util} gives the absolute level behind those ratios. At
$D = 256$ and $T \ge 8$K, FlyWheel holds 70--83\% of the BF16 peak, with a
median of 77\%, against 51--72\% for \splash{} and 22--54\% for \rpa{}.\footnote{The grid was measured at a different time from Figure~\ref{fig:softmax-kernel}, so cells shared by both differ slightly because of run-to-run variation.} At
$D = 128$ every kernel drops by roughly half, to 38--44\%, 29--39\%, and
12--30\%: halving the head dimension halves the arithmetic intensity of the
matrix products but not the softmax between them. The ordering is unchanged.

\textbf{Independence from the batch size.}
Comparing each cell's kernel throughput with the same head configuration and
length at $B = 1$ gives 616 ratios with a median deviation of 1.0\% and a 90th
percentile of 3.4\%. Only 27 exceed 5\%, and 24 of those are at $T \le 2$K, where one sequence does not fill the kernel's
2048-row query block (Appendix~\ref{app:implementation}). Throughput is thus a
function of the total token count, not of its division into sequences, which is
why $B = 1$ in \S\ref{sec:eval-softmax} represents the rest of the grid.

\section{Decode}
\label{app:decode}

FlyWheel serves decode over a paged KV cache with the same kernel design as prefill: each row accumulates against the checked static anchor of \S\ref{sec:design-softmax}, and the fragments of a staged KV block are software-pipelined, so one fragment's $\bm{P}\bm{V}$ overlaps the next fragment's $\bm{Q}\bm{K}^\top$ and exponentials. A decode step has one query row per head, so the cache read, not the matrix work, bounds it. As in \rpa{}~\cite{jiang2026ragged}, FlyWheel pads each staged token to an odd number of 32-bit word rows, so the eight sublanes of a strided load fall in different VMEM banks, and scores several KV heads through one 8-row query tile, so the softmax state of a head group fills whole registers instead of one row each. Beyond \rpa{}, FlyWheel implements ReplaySSM~\cite{liou2026replayssmcachessminputs} decode for GDN and KDA on the TPU, and supports FP8 KV caches. On v7x, FlyWheel's decode is up to 1.14$\times$ faster than \rpa{} and its speculative verify up to 1.62$\times$ faster than \brpa{}; on v6e, ReplaySSM speeds up GDN decode by up to 2.32$\times$ and 4-token verify by up to 3.72$\times$.

\textbf{BF16 decode.}
On a v7x TensorCore, FlyWheel reads the cache at 3.20--3.36~TB/s, 86--91\% of the 3.7~TB/s HBM peak, against 2.95--3.23~TB/s for \rpa{}, and is up to 1.14$\times$ faster (Table~\ref{tab:decode-bf16}). The gain is largest under MHA, where each KV head has a single query row and the shared query tile matters most: per 512-token block, FlyWheel's loop compiles to 5223 bundles, against more than 6600 for \rpa{}'s. On v6e, whose 1.6~TB/s of HBM bandwidth is less than half of v7x's (Table~\ref{tab:hardware}), the cache read bounds the step for both kernels, and they are equally fast.

\begin{table}[!ht]
\caption{BF16 decode against \rpa{} on one v7x TensorCore: 32 query heads, $D = 256$, 128-token pages, 15K cached tokens per sequence followed by 1024 decode steps. Time is ms per step.}
\label{tab:decode-bf16}
\centering
\small
\begin{tabular}{@{}llrrr@{}}
\toprule
$H_q{:}H_{kv}$ & Sequences & \rpa{} & FlyWheel & Speedup \\
\midrule
32:4  & 256 & 5.148  & 5.023  & 1.02$\times$ \\
32:4  & 128 & 2.574  & 2.519  & 1.02$\times$ \\
32:4  & 32  & 0.660  & 0.649  & 1.02$\times$ \\
32:32 & 128 & 22.426 & 19.875 & 1.13$\times$ \\
32:32 & 32  & 5.645  & 4.950  & 1.14$\times$ \\
\bottomrule
\end{tabular}
\end{table}

\textbf{Speculative verify.}
Speculative decoding verifies the draft tokens of each request in one call. FlyWheel runs a verify window of up to 16 tokens on the decode kernel and folds the window into the query heads: window token $t$ of query head $h$ becomes one more query row of $h$'s KV head, with a bottom-right causal bound. Every window token reuses the same staged K/V block, so the cache read, which bounds the step, does not grow with the window: on v6e, from $T = 1$ to $T = 16$ FlyWheel's step grows by 0--6\%, against 11--33\% for \brpa{}. \torchtpu{}~\cite{vllmtorchtpu} verifies with \brpa{}, which runs a window of $T$ tokens as a decode query of $T$ rows. \rpa{} instead sends a window to its mixed prefill path, which pads it to a full query block and takes 3.8--4.4$\times$ the time of \brpa{} at $T = 16$, so we compare with \brpa{} (Table~\ref{tab:decode-verify}). At $T = 16$, FlyWheel is 1.10--1.28$\times$ faster than \brpa{} on v6e and 1.17--1.62$\times$ on v7x, and on v7x it leads already at $T = 1$, by 1.12--1.47$\times$.

\begin{table}[!ht]
\caption{Speculative verify against \brpa{}: 32 query and 8 KV heads, $D = 128$, 128-token pages, a window of $T$ tokens per request. Time is ms per verify step, with JAX 0.11.0 and libtpu 0.0.44.}
\label{tab:decode-verify}
\centering
\small
\begin{tabular}{@{}rrrrrrrrr@{}}
\toprule
 & & & \multicolumn{3}{c}{v6e chip} & \multicolumn{3}{c}{v7x TensorCore} \\
\cmidrule(lr){4-6}\cmidrule(l){7-9}
Cached & Batch & $T$ & FlyWheel & \brpa{} & Speedup & FlyWheel & \brpa{} & Speedup \\
\midrule
2048 & 32  & 1  & 0.245 & 0.249 & 1.02$\times$ & 0.103 & 0.151 & 1.47$\times$ \\
     &     & 4  & 0.249 & 0.260 & 1.04$\times$ & 0.153 & 0.150 & 0.98$\times$ \\
     &     & 16 & 0.259 & 0.332 & 1.28$\times$ & 0.200 & 0.240 & 1.20$\times$ \\
\addlinespace
2048 & 128 & 1  & 0.982 & 0.991 & 1.01$\times$ & 0.381 & 0.442 & 1.16$\times$ \\
     &     & 4  & 0.997 & 1.020 & 1.02$\times$ & 0.567 & 0.457 & 0.81$\times$ \\
     &     & 16 & 1.032 & 1.154 & 1.12$\times$ & 0.693 & 0.812 & 1.17$\times$ \\
\addlinespace
8192 & 32  & 1  & 0.979 & 0.980 & 1.00$\times$ & 0.368 & 0.438 & 1.19$\times$ \\
     &     & 4  & 0.981 & 1.006 & 1.03$\times$ & 0.384 & 0.457 & 1.19$\times$ \\
     &     & 16 & 0.989 & 1.096 & 1.11$\times$ & 0.483 & 0.784 & 1.62$\times$ \\
\addlinespace
8192 & 128 & 1  & 3.900 & 3.872 & 0.99$\times$ & 1.429 & 1.594 & 1.12$\times$ \\
     &     & 4  & 3.865 & 3.934 & 1.02$\times$ & 1.481 & 1.693 & 1.14$\times$ \\
     &     & 16 & 3.904 & 4.302 & 1.10$\times$ & 1.858 & 2.999 & 1.61$\times$ \\
\bottomrule
\end{tabular}
\end{table}

\textbf{FP8 KV cache.}
An FP8 cache stores K and V in \texttt{float8\_e4m3fn} or \texttt{float8\_e5m2} with one descale per tensor, as in FlashAttention-3~\cite{shah2024flashattention}. A 32-bit word holds the (K, K, V, V) bytes of a head pair, so the cache keeps the BF16 interleave at head-pair granularity. The MXU multiplies the BF16 query by the FP8 keys and values directly; the K descale folds into the query's scale and the V descale into the softmax normalization. Both descales are runtime values, so layers with different scales share one compiled kernel. On v6e, FP8 halves the step of FlyWheel's BF16 decode, 1.98$\times$ and 1.97$\times$ faster at batch 32 and 128 (0.979 to 0.495~ms and 3.900 to 1.979~ms), and matches \rpa{}'s FP8 decode (Table~\ref{tab:decode-fp8}).

\begin{table}[!ht]
\caption{FP8 decode against \rpa{} on one v6e chip: 32 query and 8 KV heads, $D = 128$, 8K cached tokens per sequence. Time is ms per call.}
\label{tab:decode-fp8}
\centering
\small
\begin{tabular}{@{}rrrr@{}}
\toprule
Batch & FlyWheel e4m3fn & FlyWheel e5m2 & \rpa{} e4m3fn \\
\midrule
32  & 0.495 & 0.494 & 0.489 \\
128 & 1.979 & 1.973 & 1.954 \\
\bottomrule
\end{tabular}
\end{table}

\textbf{Linear attention decode with ReplaySSM.}
ReplaySSM~\cite{liou2026replayssmcachessminputs} keeps each sequence's recurrent state as a checkpoint plus the inputs of up to $W$ recent tokens. A decode step computes its output from the checkpoint and the buffered inputs without materializing the state, and writes the checkpoint only when the buffer overflows; a verify step commits the accepted tokens by advancing the buffer length. FlyWheel fuses this step with the causal conv1d. It folds the buffers of a tile's overflowing sequences into new checkpoints with one FP32 matmul issued halfway through the tile, so the checkpoint DMAs overlap the other sequences' compute. It stores the FP32 checkpoint as two 16-bit planes: a decode step reads only the high plane, the BF16 value its matmul consumes, and a flush rebuilds the FP32 value bit for bit. Table~\ref{tab:decode-replay} compares with FlyWheel's conventional decode kernel, which reads and writes the full state at every step and runs at the speed of \gdnv{}. ReplaySSM is 1.77--2.32$\times$ faster on decode and 3.26--3.72$\times$ on 4-token verify for GDN, and 1.89--2.17$\times$ and 2.91--3.43$\times$ for KDA. The gain is largest on verify: a conventional kernel writes a recurrent state for every window position, whereas ReplaySSM commits the window by advancing the buffer length.

\begin{table}[!ht]
\caption{ReplaySSM decode on one v6e chip: 16 query/key heads, 32 value heads, $d_k = d_v = 128$, $W = 16$, BF16 buffered inputs. Conv.\ is FlyWheel's conventional decode kernel. Time is ms per call; verify-4 checks 4 tokens per sequence.}
\label{tab:decode-replay}
\centering
\small
\begin{tabular}{@{}lrrrrrrr@{}}
\toprule
 & & \multicolumn{3}{c}{GDN} & \multicolumn{3}{c}{KDA} \\
\cmidrule(lr){3-5}\cmidrule(l){6-8}
Step & Batch & Conv. & ReplaySSM & Speedup & Conv. & ReplaySSM & Speedup \\
\midrule
Decode   & 16  & 0.101 & 0.057 & 1.77$\times$ & 0.105 & 0.055 & 1.91$\times$ \\
         & 64  & 0.264 & 0.114 & 2.32$\times$ & 0.267 & 0.123 & 2.17$\times$ \\
         & 128 & 0.478 & 0.207 & 2.31$\times$ & 0.478 & 0.249 & 1.92$\times$ \\
         & 256 & 0.940 & 0.447 & 2.10$\times$ & 0.944 & 0.500 & 1.89$\times$ \\
\addlinespace
Verify-4 & 16  & 0.189 & 0.058 & 3.26$\times$ & 0.205 & 0.065 & 3.15$\times$ \\
         & 64  & 0.667 & 0.181 & 3.69$\times$ & 0.670 & 0.230 & 2.91$\times$ \\
         & 128 & 1.324 & 0.357 & 3.71$\times$ & 1.326 & 0.387 & 3.43$\times$ \\
         & 256 & 2.657 & 0.715 & 3.72$\times$ & 2.675 & 0.779 & 3.43$\times$ \\
\bottomrule
\end{tabular}
\end{table}

\section{Additional Implementation Details}
\label{app:implementation}

\textbf{Softmax fragments.}
The implementation uses \texttt{exp2}, folding $\log_2 e$ and the attention scaling factor into $\bm{Q}$ before the fragment loop. A $2048\times2048$ block contains 32 fragments of 256 query rows and 512 key rows. The loop is unrolled into one compiler basic block; when a block contains multiple heads, the final fragment of one head overlaps the first of the next. The previous fragment's weights remain live until its output product completes, and the final output product drains the pipeline.

\textbf{GDN matrix execution.}
The kernel transposes $\bm{K}$ once on the XLU and reuses the result in both stacked products. The hierarchical inverse uses two products for the base blocks and three per merge. On a repeated-key test at $C=64$, its maximum absolute error relative to FP64 is $9\times10^{-3}$. Convolution and normalization complete before the cross-chunk pipeline, whose independent work is exposed within one compiler basic block.

\textbf{Convolution and vector operations.}
Unused trailing rows of the tap-scaled copies in $\bm{W}$ hold the preceding chunk's halo. For $k=4$ and $C=64$, the convolution's contraction dimension is 256. The selector is labeled $\bm{S}$ in Figure~\ref{fig:vector}b and is distinct from the recurrent state. The remaining vector operations use
\[
\begin{aligned}
\mathrm{SiLU}(x)&=\tfrac{x}{2}(1+\tanh(x/2)),\\
\frac{\bm{x}}{\sqrt{\|\bm{x}\|_2^2+\epsilon}}&=\bm{x}\,\mathrm{rsqrt}\!\left(\textstyle\sum_j x_j^2+\epsilon\right).
\end{aligned}
\]
Here $\epsilon=10^{-6}$.
Table~\ref{tab:gdn-ablation} measures their combined benefit with the inverse and convolution reformulations.

\textbf{Kimi Delta Attention.}
For KDA's per-channel decay, FlyWheel factors decay across intermediate block boundaries and incorporates it into masked matrix products. The resulting inverse uses ten matrix multiplications and follows the same principle of overlapping chunk-local computation with state-dependent work from another chunk.

\end{document}